\documentclass{article}

\usepackage{microtype}
\usepackage{graphicx}
\usepackage{multirow}
\usepackage{booktabs} 

\usepackage{hyperref}

\usepackage[accepted]{icml2024}

\usepackage{amsmath}
\usepackage{amssymb}
\usepackage{mathtools}
\usepackage{amsthm}

\usepackage[capitalize,noabbrev]{cleveref}

\theoremstyle{plain}
\newtheorem{theorem}{Theorem}[section]
\newtheorem{proposition}[theorem]{Proposition}
\newtheorem{lemma}[theorem]{Lemma}

\theoremstyle{definition}

\newtheorem{assumption}[theorem]{Assumption}
\theoremstyle{remark}

\usepackage[textsize=tiny]{todonotes}

\usepackage{color,soul}
\usepackage{mathnotation}

\newcommand*\diff{\mathop{}\!\mathrm{d}}
\usepackage{subcaption}
\usepackage{algorithmic}
\usepackage{algorithm}
\usepackage{subcaption}

\icmltitlerunning{Path-Guided Particle-based Sampling}

\begin{document}

\twocolumn[
\icmltitle{Path-Guided Particle-based Sampling}



\icmlsetsymbol{equal}{*}

\begin{icmlauthorlist}
\icmlauthor{Mingzhou Fan}{equal,1}
\icmlauthor{Ruida Zhou}{equal,2}
\icmlauthor{Chao Tian}{1}
\icmlauthor{Xiaoning Qian}{1,3,4}
\end{icmlauthorlist}

\icmlaffiliation{1}{Department of Electrical \& Computer Engineering, Texas A\&M University, College Station, Texas, USA}
\icmlaffiliation{2}{Department of Electrical and Computer Engineering, University of California, Los Angeles, CA, USA}
\icmlaffiliation{3}{Department of Computer Science and Engineering, Texas A\&M University, College Station, TX, USA}
\icmlaffiliation{4}{Computational Science Initiative, Brookhaven National Laboratory, Upton, NY, USA}

\icmlcorrespondingauthor{Mingzhou Fan}{mzfan@tamu.edu}
\icmlcorrespondingauthor{Xiaoning Qian}{xqian@tamu.edu}

\icmlkeywords{Machine Learning, ICML}

\vskip 0.3in
]



\printAffiliationsAndNotice{\icmlEqualContribution} 

\begin{abstract}
Particle-based Bayesian inference methods by sampling from a partition-free target (posterior) distribution, e.g., Stein variational gradient descent (SVGD), have attracted significant attention. 
We propose a path-guided particle-based sampling~(PGPS) method based on a novel Log-weighted Shrinkage (LwS) density path linking an initial distribution to the target distribution. 
We propose to utilize a Neural network to learn a vector field motivated by the Fokker-Planck equation of the designed density path. 
Particles, initiated from the initial distribution, evolve according to the ordinary differential equation defined by the vector field. The distribution of these particles is guided along a density path from the initial distribution to the target distribution.
The proposed LwS density path allows for an efficient search of modes of the target distribution while canonical methods fail. We theoretically analyze the Wasserstein distance of the distribution of the PGPS-generated samples and the target distribution due to approximation and discretization errors. Practically, the proposed PGPS-LwS method demonstrates higher Bayesian inference accuracy and better calibration ability in experiments conducted on both synthetic and real-world Bayesian learning tasks, compared to baselines, such as SVGD and Langevin dynamics, etc. 
\end{abstract}


\section{Introduction}
Bayesian learning is a powerful approach for distribution-based model predictions, naturally equipped with uncertainty quantification and calibration powers \citep{pml1Book}. The key of Bayesian learning -- computing the posterior by Bayes' rule, however, is well-known to be challenging due to the intractable partition function (a.k.a. the normalizing constant)~\citep{andrieu2003introduction}. 

To circumvent this difficulty, approaches based on sampling according to the (target) posterior distribution without computing the partition function have been considered;  e.g., Markov Chain Monte-Carlo (MCMC) sampling~\citep{andrieu2003introduction} and its gradient-based variants (e.g., Langevin dynamics) generate samples (or \emph{particles}) that follow the target distribution asymptotically using a partition-free function. Such particle-based Bayesian inference methods, which essentially transform a set of initial samples/particles along certain dynamics (e.g., an ordinary differential equation (ODE) or a stochastic differential equation (SDE)) governed by a vector field, have witnessed great successes~\citep{liu2017stein}. 
Most of these methods, e.g. Stein variational gradient descent (SVGD)~\citep{liu2016stein} and preconditioned functional gradient flow (PFG)~\cite{dong2022particle}, fall into the category of \emph{gradient-flow} particle-based sampling, where the vector field is a gradient function of the Kullback-Leibler (KL) divergence 
of the current distribution to the target distribution, such that the dynamics would drive the particles to the minimum of KL-divergence solution, i.e., the target distribution. 

Although \emph{gradient-flow} particle-based sampling methods are shown to be flexible and efficient in some applications~\citep{dong2022particle}, they may not achieve the ideal Bayesian inference performance due to not effectively capturing the posterior distribution. 
Specifically, as a realization of the KL-Wasserstein gradient-flow method, Langevin Dynamic (LD) is known to suffer from slow mixing, and in turn tends to result in mode missing or misplaced mode weights~\citep{song2019generative}. It is believed that the posterior for complicated models, especially Bayesian Neural Networks~(BNNs)~\citep{goan2020bayesian}, contain multiple modes of different weights, and mode missing would impact its generalization, uncertainty quantification, and calibration abilities. 
More detailed discussions can be found in Sections~\ref{sec:pathsel} and ~\ref{sec:Gaussian}. 

In this work, we propose a new family of Bayesian inference methods
termed \textbf{Path-Guided Particle-based Sampling~(PGPS)}. The particles follow an ODE defined by a learned vector field, so that the distribution of the particles is directed by a carefully designed \emph{partition-free} path connecting the initial and the target distributions, instead of evolving along the direction that minimizes some functional. 
The performance of the PGPS approach is clearly determined by the path being followed, and we propose to rely on a \emph{Log-weighted Shrinkage} path that is more efficient and accurate. The intuition for this choice is that logarithmic weights admit linear mixture of the score function and the shrinkage allows effective coverage of the target distribution along the path. 



The contributions of this work are threefold:
\begin{enumerate}
\item We propose PGPS as a novel framework of flow-based sampling methods 
and derive a tractable criterion for any differentiable partition-free path in Proposition \ref{prop:equality};
\item We theoretically show that the Wasserstein distance between the target distribution and the PGPS generated distribution following the NN-learned vector field with approximation error $\delta$ and discretization error by step-size $h$ is bounded by $\Oc(\delta) + \Oc(\sqrt{h})$ in Theorem \ref{thm:combine};
\item We experimentally verify the superior performance of the proposed approach over the state-of-the-art benchmarks, in terms of the sampling quality of faster mode seeking and more accurate weight estimating, and the inference quality with \emph{higher testing accuracy} and \emph{stronger calibration ability} in Bayesian inference, in Section \ref{sec:experiment}.
\end{enumerate}

\section{Background}
Given an inference model parameterized by parameters $\rvx$, e.g., a neural network with parameters $\rvx$, Bayesian inference updates the distribution of the parameters by Bayes' theorem, and performs statistical inference according to the posterior distribution. Specifically, suppose parameters $\rvx \in \Rb^d$ has prior density\footnote{We assume density of the parameters (random variables) exists and do not differentiate their distribution and density.} $p_0(\rvx)$, and given a data set $\mathcal{D}$, the posterior $p^*(\rvx)$ is updated by $p^*(\rvx) = \frac{\hat{p}(\rvx)}{Z}$ with $\hat{p}(\rvx) = p_0(\rvx)L(\mathcal{D}|\rvx)$, where $L(\mathcal{D}|\rvx)$ is the likelihood function of the data $\mathcal{D}$ and $Z = \int p_0(\rvx)L(\Dc | \rvx) \diff \rvx$ is the partition function. The partition function $Z$ is usually computationally intractable. Many inference methods, including broadly applied Monte Carlo methods~\citep{liu2001monte}, have been proposed to (approximately) draw samples from the posterior/target distribution $p^*(\rvx)$ using the more tractable partition-free function $\hat{p}(\rvx)$.

Particle-based (particularly flow-based) Bayesian inference methods direct a set of random samples/particles $\{\rvx_{0}^{(i)} \}_{i=1}^n \subset \mathbb{R}^d$ drawn i.i.d. from an initial distribution $p_0$ (e.g., the prior or other distributions from which samples can be drawn directly) along certain ODE dynamics
\begin{equation*}
\frac{\diff \rvx_t}{\diff t} = \boldsymbol{\phi}_t(\rvx_t), \quad \rvx_0 \sim p_0,
\end{equation*}
defined by a vector field $\boldsymbol{\phi}_t: \mathbb{R}^d \rightarrow \mathbb{R}^d$. 
The corresponding evolution of the density functions is characterized by the {continuity} equation ~\citep{jordan1998variational}:
\begin{equation}\label{eq:continuity}
    \frac{\partial}{\partial t} p_t(\rvx) = - \nabla \cdot (p_t(\rvx) \boldsymbol{\phi}_t(\rvx)),
\end{equation}
where $p_t(\rvx)$ denotes the density of $\rvx_t$, $\nabla$ is the vector differential operator w.r.t. $\rvx$ (we omit $\rvx$ for simplicity throughout the paper), and $\nabla \cdot \vf$ denotes the divergence of the vector function $\vf$. 

The critical point of the particle-flow-based methods is the design of the vector field $\boldsymbol{\phi}_t$. A typical choice is the gradient of some objective function under a certain metric, and the dynamic is thus a \emph{gradient flow}. An example of the gradient-flow particle-based method is 
the Wasserstein gradient flow~\citep{ambrosio2005gradient}, which has drawn considerable interest. It is motivated by minimizing a functional $L(p_t) \in \Rb$ in the Wasserstein space, which is a space of distributions equipped with the Wasserstein metric
\begin{equation}
W_q(p_1, p_2) = \left(\inf_{\gamma \in \Gamma(p_1, p_2)} \mathbb{E}_{(\rvx_1, \rvx_2) \sim \gamma}\|\rvx_1 - \rvx_2\|_q^{q} \right)^{1/q}, \notag
\end{equation}
where $\Gamma(p_1, p_2)$ is the set of all the coupling of $p_1$ and $p_2$. 
When the functional is the KL divergence $\KL(p_t\|p^*) = \mathbb{E}_{\rvx_t\sim p_t}[-\ln p^*(\rvx_t) + \ln p_t(\rvx_t)]$ and under the 2-Wasserstein metric, i.e. $q=2$, the resulting gradient has a closed form \begin{equation}\label{eq:wgf}
\boldsymbol{\phi}_t(\rvx) = \nabla \ln p^*(\rvx) - \nabla \ln p_t(\rvx).
\end{equation} 
Under mild assumptions, the gradient flow converges to the optimal solution, i.e., $\lim_{t \rightarrow \infty} p_t = p^*$, which implies that with sufficiently large $t$, $\rvx_t$ approximately follows the target distribution $p^*$. 
Computing $\nabla \ln p_t(\rvx)$ in~\cref{eq:wgf} is however not feasible in most practical cases. Methods such as learning the current density $p_t(\rvx)$~\citep{wang2022projected}, or transforming the problem of finding $\boldsymbol{\phi}_t(\rvx)$ to a tractable learning/optimization problem~\citep{dong2022particle, di2021neural} by a swarm of particles at step $t$, have been developed to implement the gradient flow. The vector field learning in our work is also based on a swarm of particles in the same manner, though the training loss function and the desired vector field are significantly different. 

Evidently, the dynamics of the particles are not unique given the evolution of the distributions. Langevin dynamics (LD) $\diff \rvx_t = \nabla \ln \hat{p}(\rvx_t) \diff t + \sqrt{2} \diff \mathbf{B}_t$, where $\mathbf{B}_t$ is a standard Brownian motion, is a realization of the KL Wasserstein gradient flow \cite{jordan1998variational}. In other words, its distribution satisfies the same Fokker-Planck equation as that of the ODE with vector field $\boldsymbol{\phi}_t(\rvx)$ in \eqref{eq:wgf}. LD and its variants, such as Metropolis-adjusted Langevin Algorithm~( {MALA or LMC}) and Stochastic Gradient Langevin Dynamics~(SGLD), have been shown to be effective since they do not require learning $\nabla \ln p_t(\rvx)$ as in \eqref{eq:wgf}. 
However, these methods lack a stopping criterion due to their stochastic nature~\citep{dong2022particle},  and can suffer from slow convergence for some target distributions. 

\subsection{Motivation of the Proposed PGPS Method}
\begin{figure}
     \centering
     \begin{subfigure}{.49\linewidth}
         \includegraphics[width=\linewidth]{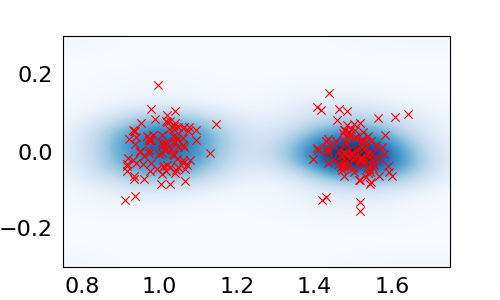}
         \caption{Target distributed samples}
         \label{fig:LMC_target}
     \end{subfigure}
     \begin{subfigure}{.49\linewidth}
         \includegraphics[width=\linewidth]{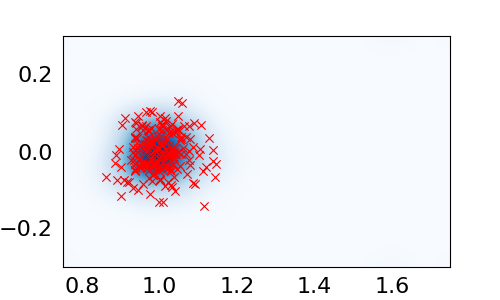}
         \caption{LD for 100 iterations}
         \label{fig:100LMC}
     \end{subfigure}
     \begin{subfigure}{.49\linewidth}
         \includegraphics[width=\linewidth]{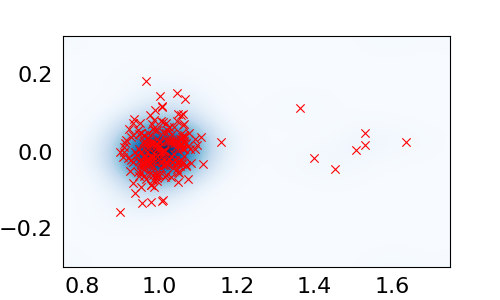}
         \caption{LD for 4,000 iterations}
         \label{fig:1000LMC}
     \end{subfigure}
     \begin{subfigure}{.49\linewidth}
         \includegraphics[width=\linewidth]{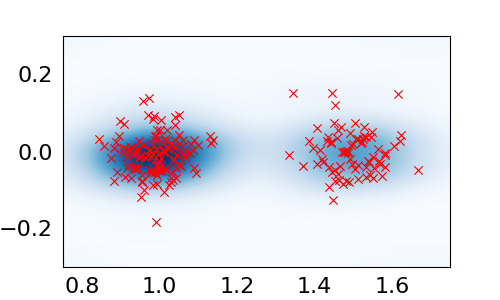}
         \caption{PGPS for 650 iterations}
         \label{fig:pggf}
     \end{subfigure}\vspace{-0mm}
     \caption{An illustration of the effectiveness of PGPS over LD in handling mode-missing. 
     }
     \vspace{-0.6cm}
     \label{fig:two_Gaussian}
\end{figure}

We first pinpoint the cause of the slow convergence of KL Wasserstein gradient flow (e.g., LD), and provide the intuition for the proposed method as a remedy. 
Consider the experiment setup with a target distribution being a mixture of two Gaussian distributions, as shown in Figure~\ref{fig:two_Gaussian} (a). Taking a zero mean isometric Gaussian as initial distribution, the convergence of LD to the target distribution is extremely slow as shown in Figure~\ref{fig:two_Gaussian} (b-c), where the particles ``stuck" at the left-hand-side mode of the Gaussian mixture and it takes many iterations to reach the right-hand-side mode. The reason for this behavior is  {that} LD and similar gradient-flow-based methods rely heavily on the target distribution, which is an asymptotic target. Such an asymptotic target does not reflect the short-term need to escape from the current mode; i.e., the convergence to the target distribution can be extremely slow. To solve this issue, we propose PGPS which specifies a density evolution path directly connecting the initial and target distribution, and let the distribution of the particles evolve along such a path. At each time step, a short-term intermediate target on the path is set for the particles; more details are given in Section \ref{sec:PGPS}. As shown in Figure~\ref{fig:two_Gaussian} (d), PGPS indeed finds both modes and converges to the target distribution with considerably fewer iterations.




In the  {following}, we denote the unnormalized density function with a hat as $(\hat{\cdot})$, i.e. $\hat{p}_t(\rvx) \propto p_t(\rvx)$ with $\int p_t(\rvx) \diff x = 1$ but $\int \hat{p}_t(\rvx) \diff \rvx$ being an unknown positive number.

\subsection{Related Works}\label{sec:related}

Gradient-flow particle-based sampling usually aims at finding tractable estimations for the KL-gradient flows in the Wasserstein space. One track of works relies on the universal approximation theorem of neural networks~\citep{hornik1989multilayer} to approximate the gradient-flow and maximize certain discrepancies~\citep{di2021neural, grathwohl2020learning, hu2018stein, dong2022particle}{, among which preconditioned functional gradient flow (PFG)~\cite{dong2022particle} was proposed to learn the Wasserstein gradient by a neural network with preconditioning for better approximation. Probability flow ODE~\citep{maoutsa2020interacting} can also be applied to learn the Wasserstein gradient flow aiming at learning the vector field for a given probability flow.} Aside from focusing on the flow approximation, works focusing on the discretization adopt the Jordan, Kinderlehrer, and Otto~(JKO) scheme~\cite{jordan1998variational}, aiming at finding a JKO operator that minimizes the target functional as well as the movement of the particles in each step, has also achieved good performance in arbitrary gradient flow other than KL-gradient flow estimation tasks~\citep{alvarez2021optimizing, mokrov2021large}.

The Stein Variational Gradient Descent~(SVGD)~\citep{liu2016stein} can be viewed as a specific type of gradient flow w.r.t. the KL-divergence under a metric induced by Stein operator, i.e., approximating the gradient by a kernel function~\citep{liu2017stein}. {It inspires later works on kernel methods following different flows, e.g., Fisher–Rao Flow~\citep{maurais2024sampling} and the flow introduced by minimizing first and second moments~\citep{wang2024measure}.} However, the curse of dimensionality for the kernel-based methods leads to the particle collapse in SVGD~\citep{ba2021understanding}, i.e., variance collapse. Projecting the inference space to a lower dimension can naturally avoid high-dimensional variational inference~\citep{chen2020projected, gong2020sliced, liu2022grassmann}. 

{
Another area of related work is the annealing-based methods, e.g., parallel tempering~\citep{earl2005parallel}, annealed importance sampling~\citep{neal2001annealed}, and sequential Monte Carlo~\cite{doucet2001introduction}. Annealing-based methods utilize intermediate distributions, usually following a log-weighted schedule where the weights are usually interpreted as temperature, to help achieve better performance. Utilizing intermediate distributions (path) has witnessed benefits in both Monte-Carlo estimators~\citep{grosse2013annealing, chehab2024provable} and sampling tasks~\citep{heng2020controlled}.

Learning vector fields to update the particles has been broadly adopted in generative models that consider a different task aiming at generating new samples based on existing ones. The backward process of diffusion models~\cite{song2020score, albergo2023stochastic} is indeed learning the vector fields that can drive the particles inverting the path introduced by the forward process. Flow matching~\citep{lipman2022flow}, built on Continuous Normalizing Flows~(CNF)~\citep{9089305} to learn a vector field following some specifically designed path, has demonstrated its empirical effectiveness followed by justification from~\citet{benton2023error} theoretically.}

In this work, we use SVGD, PFG and LD as benchmarks 
and defer more detailed discussions on related works to~\cref{app:related}.

\section{Path-Guided Particle-based Sampling \label{sec:PGPS}}
We propose Path-Guided Particle-based Sampling (PGPS) methods based on a continuous density path linking initial distribution $p_0$ to target distribution $p_1 = p^*$, while only accessing the partition-free version of the target distribution $\hat{p}_1 = \hat{p}$.  {Compared to the annealing methods that also utilize intermediate distributions (path), PGPS learns a vector field that would drive the particles to the next intermediate distribution based on a predefined path in each step, which has not been studied previously to the best of our knowledge.} In this section, we first derive a condition for viable guiding paths and present a novel class of log-weighted shrinkage paths. We then propose a learning algorithm to effectively approximate the path-guided flow.

Given a partition-free density process $\{\hat{p}_t \}_{t \in [0, 1]}$ and its normalized densities $\{p_{t}\}_{t \in [0, 1]}$, with $\hat{p}_0 = p_0$ being the initial distribution and $p_1$ being the target, assume that $\frac{\partial}{\partial t} \hat{p}_t$ and $\nabla_\vx \hat{p}_t(\vx)$ exist for any $t\in [0, 1]$ and $\vx$ on the support. We wish to construct a vector field $\boldsymbol{\phi}_t : \mathbb{R}^d \rightarrow \mathbb{R}^d$ such that the process
\begin{equation}\label{eq:evolve}
    \frac{\mathrm{d} \rvx_t}{\mathrm{d} t} = \boldsymbol{\phi}_t( \rvx_t ), \quad \rvx_0 \sim p_0
\end{equation}
satisfies $\rvx_t \sim p_t$ for any $t \in [0, 1]$. 
The following proposition establishes that determining $\boldsymbol{\phi}_t(\vx)$ does not require the partition function. 

\begin{proposition}\label{prop:equality}
    For a given partition-free density path $\{\hat{p}_t\}$, the gradient flow guided by the vector field $\boldsymbol{\phi}_t(\rvx)$ following the continuity equation~\eqref{eq:continuity} satisfies:
    \begin{equation}\label{eq:pggf}
    r(\vx, \boldsymbol{\phi}_t) - \mathbb{E}_{\rvx \sim p_t}\left[\frac{\partial \ln\hat{p}_t(\rvx)}{\partial t}\right]=0,
    \end{equation}
    where $r(\vx, \boldsymbol{\phi}_t) = \frac{\partial \ln\hat{p}_t(\vx)}{\partial t}  + (\nabla \ln{\hat{p}_t}(\vx) + \nabla) \cdot \boldsymbol{\phi}_t(\vx)$. 
\end{proposition}

The proof of~\cref{prop:equality} can be found in~\cref{app:proof}.


Proposition~\ref{prop:equality} indicates that once a vector field $\boldsymbol{\phi}_t$ satisfying Equation~(\ref{eq:pggf}) is obtained, we can generate samples following the distribution on the density path $\{p_t\}$ when particles evolve according to the vector field in Equation \eqref{eq:evolve}. 
Furthermore, note that Equation \eqref{eq:pggf} is free of the intractable partition function, and we can thus learn the vector field $\boldsymbol{\phi}_t(\vx)$ by approximating it via a neural network  {that} solves for Equation (\ref{eq:pggf}). 




\subsection{Selection of Path}\label{sec:pathsel}

One of the most important components of the proposed approach is the selection of partition-free guiding path $\{\hat{p}_t\}_{t \in [0,1]}$. Although any reasonable path linking the initial and target distributions is valid to direct particles according to Equation \eqref{eq:evolve} as long as the corresponding vector field follows the condition in Proposition \ref{prop:equality}, certain paths that are more robust against democratization and more tractable for training are preferred and may have better performance in practice.


We propose a class of \emph{Log-weighted Shrinkage} paths $\{\hat{p}_t^{\text{LwS}}\}$ as follows
\begin{equation}
\begin{aligned}
\footnotesize \ln \hat{p}_t^{\text{LwS}}(\vx) &:=({1-t}) \ln p_0\left((1-\alpha t)\vx\right)  \\
&\qquad+ t \ln \hat{p}_1 \left(\frac{\vx}{\beta +(1 - \beta) t}\right),
\end{aligned}
\label{eq:logspath}
\end{equation}
where $\alpha \in [0, 1]$ and $\beta \in (0, 1]$ are controlling parameters.


It is straightforward to check that LwS paths are valid with $\ln \hat{p}_0^{\text{LwS}}(\vx) = \ln p_0(\vx)$ and $\ln \hat{p}_1^{\text{LwS}}(\vx) = \ln \hat{p}_1(\vx)$. Moreoever, $\frac{\partial}{\partial t} \ln \hat{p}_t^{\text{LwS}}$ and $\nabla \ln \hat{p}_t^{\text{LwS}}$ both exist, when $\nabla \ln \hat{p}_1$ and $\nabla \ln p_0$ exist; see~\cref{app:gradient}. 

\begin{figure*}[t]
\includegraphics[width=\linewidth]{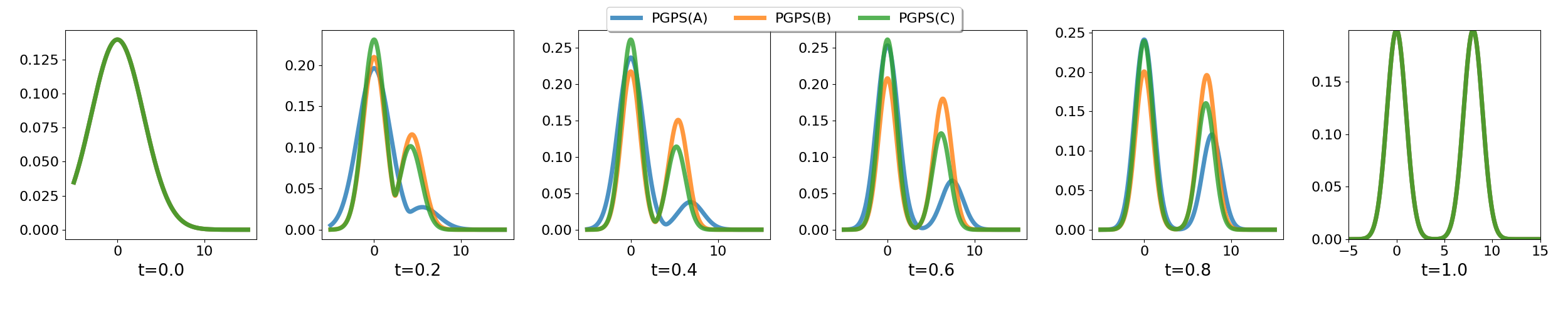}\vspace{-5mm}
    \caption{Different \emph{Log-weighted Shrinkage paths} from the initial (left) to target (right) distribution with different hyper-parameters. (A):$\alpha=0, \beta=1$ (blue); (B):$\alpha=1, \beta=0.5$ (orange); (C):$\alpha=0.2, \beta=0.5$ (green). \label{fig:paths}}
\end{figure*}

As its name suggested, LwS paths \eqref{eq:logspath} have two components -- Log-weights and Shrinkage. The log-weights enable representing the log-distribution on the path by a linear mixture of the log-initial-distribution and log-target-distribution terms in Equation \eqref{eq:logspath} weighted by $(1-t)$ and $t$. The linear mixture allows efficient computation of $r(\vx, \boldsymbol{\phi}_t)$ in Proposition \ref{prop:equality} when training $\boldsymbol{\phi}_t$ by a neural network. The Shrinkage operates on the initial-distribution term by $\alpha$ and the target-distribution term by $\beta$ in Equation \eqref{eq:logspath}. The first term spreads the initial distribution by a factor $1/(1-\alpha t)$ to cover larger ranges as the factor increases along $t$; and the second term shrinks the target distribution $\hat{p}_1$ towards zero (i.e., the distribution $\hat{p}_1 (\frac{\vx}{\beta +(1 - \beta) t})$ is thinner than $\hat{p}_1 (\vx)$) by a factor $\beta + (1-\beta) t$. Since a typical choice of $p_0$ is zero-mean Gaussian, the shrinkage allows better coverage of the target distribution, and the coverage enables better mode seeking. It is illustrated in  Figure~\ref{fig:paths} for different choices of the hyperparameters $\alpha, \beta$. We can observe that with appropriate choices of hyperparameters (e.g., (B) and (C)), the right mode of the target distribution is detected at an early stage (e.g., $t = 0.2$) compared to the log-weighted path without shrinkage (e.g., (A)).  {We further discuss the influence of the choice of the hyperparameters in Appendix~\ref{app:hyper}.}

\subsection{Learning Vector Field $\boldsymbol{\phi}_t(\vx)$ \label{sec:learning-vector-field}}
Given a viable path $\{p_t\}$, we aim to find a corresponding vector field $\boldsymbol{\phi}_t(\vx)$ as in Proposition \ref{prop:equality} to direct the particles as in Equation \eqref{eq:evolve}. However, solving Equation \eqref{eq:pggf} for $\boldsymbol{\phi}_t(\vx)$ in closed form is intractable. We use a parameterized vector field model $\boldsymbol{\phi}_t^\theta(\vx) \in \Rb^d$ -- a neural network parameterized by $\theta$ -- to approximately solve for Equation \eqref{eq:pggf}. 

Specifically, at each time step $t$ starting with $t = 0$, we have $N$ particles $\{ \vx_{t, (i)}\}_{i = 1,\ldots, N}$ and minimize the training loss
\begin{equation}\label{eq:pggfloss}\footnotesize
L_t(\theta) =\sum_{i = 1 \dots N}
\left|r(\vx_{t, (i)}, \boldsymbol{\phi}_t^\theta) - \frac{1}{N}\sum_{j = 1 \dots N}\frac{\partial \ln\hat{p}_t(\vx_{t, (j)})}{\partial t} \right|^2
\end{equation}
resembling the squared value of the LHS of Equation \eqref{eq:pggf}. When particles $\{ \vx_{t, (i)}\}$ following distribution $p_t$, $\frac{1}{N}\sum_{j = 1 \dots N}\frac{\partial \ln\hat{p}_t(\vx_{t, (j)})}{\partial t}$ is an unbiased estimate of $\mathbb{E}_{\rvx \sim p_t}\left[\frac{\partial \ln\hat{p}_t(\rvx)}{\partial t}\right]$. 

The training algorithm is presented in Algorithm~\ref{alg:train}, where the loss \eqref{eq:pggfloss} is minimized by gradient descent. It is an iterative algorithm starting from time step $t = 0$ and is increased by $\Delta t$ after the training for time $t$. The time step increment $\Delta t$ is adaptively determined by Algorithm~\ref{alg:timestep}, which leads to a smaller increment for larger vector field $\boldsymbol{\phi}_t^{\theta}$ to control the movement of the particles. Since we have an intermediate target distribution $\hat{p}_t$ on the path to follow, an optional Langevin adjustment (Langevin dynamics w.r.t. the intermediate target $\hat{p}_t$) in Algorithm \ref{alg:LD} can be applied to adjust the particles' distribution closer to $p_t$ to reduce the biasedness in the loss function \eqref{eq:pggfloss}. We further discuss the Langevin Adjustment in the experiment section.




\begin{algorithm}[]
   \caption{Adaptive Time Step}
   \label{alg:timestep}
\begin{algorithmic}
   \STATE {\bfseries Input:} Time $t$, Current particles $\{\vx_{t, (i)}\}_{i=1\dots N}$, Flow $\boldsymbol{\phi}_t^\theta(\vx)$, Particle step-size $\psi$, Maximum time step $\Delta t'$;
   \STATE $\Delta t \gets (N \psi) / \sum_{i = 1 \dots N} \left\|\boldsymbol{\phi}_t^\theta(\vx_{t, (i)})\right\|$;
   \STATE $\Delta t \gets \min\{\Delta t, 1-t, \Delta t'\}$;
   \STATE {\bfseries Output:} Time step $\Delta t$;
\end{algorithmic}
\end{algorithm}

\begin{algorithm}[]
   \caption{Langevin Adjustment}
   \label{alg:LD}
\begin{algorithmic}
   \STATE \texttt{ // Langevin dynamics //}
   \STATE {\bfseries Input:} Particles $\{\vx_{(i)}\}_{i=1\dots N}$, density $\hat{p}$; \STATE {\bfseries Coefficients:} Adjustment step-size $\delta$, LD steps $M'$;
   \FOR{$k = 1 \dots M'$}
   \STATE Sample $\{\xi_{(i)}^k\} \sim \mathcal{N}(0, I)$;
    \STATE Adjust $\{\vx_{(i)}\} \gets \{\vx_{(i)} + \delta \nabla \ln \hat{p}(\vx_{(i)}) + \sqrt{2\delta} \xi_{(i)}^k\}$;
    \ENDFOR
   \STATE {\bfseries Output:} Adjusted $\{\vx_{(i)}\}$;
\end{algorithmic}
\end{algorithm}

\paragraph{Training-free deployment of PGPS} 
Many efficient algorithms such as LD or SVGD, are training-free, i.e., learning is not required during the evolution of the particles. We can also implement PGPS in a training-free manner, where at each time step $t$ without training a neural network we update the particles by Langevin adjustment solely. In other words, we iteratively apply Langevin dynamics for sampling from an intermediate target distribution $\hat{p}_t$. A similar approach has been proposed under the name Annealed Langevin Dynamics~(ALD)~\cite{song2019generative}, where a path is given by changing the temperature of the target distribution. In Section~\ref{app:ALD}, we experimentally compare the standard PGPS and the training-free PGPS and demonstrate the benefits of learning the vector field.



\section{Theoretical Analysis}
In this section, we study the distribution of the PGPS-generated particles compared to the target distribution. Note that the target distribution $p^* \propto \hat{p}$ equals to $p_{\rvx_1}$, where $\rvx_1 = \rvx_0 + \int_0^1 \boldsymbol{\phi}_t(\rvx_t) \diff t$ with $\rvx_0 \sim p_0$ by Proposition~\ref{prop:equality}. The PGPS method without Langevin adjustment simulates the integration by 
\begin{align}
    \hat{\rvx}_{th + h}^\theta = \hat{\rvx}_{th}^\theta + \boldsymbol{\phi}_{nh}^\theta(\hat{\rvx}_{th}^\theta), \quad t = 0,\ldots, n-1, \label{eq:descrete-PGPS}
\end{align}
where $h = 1/n$ is the step size for some $n \in \Nb$ capturing the discretization error, and $\hat{\rvx}_{0}^\theta \sim p_0$.

We analyze the performance of PGPS using the 2-Wasserstein distance between the generated distribution $p_{\hat{\rvx}_1}$ and the target distribution $p_{\rvx_1}$ under the approximation error {  $\delta^2 := \int_{0}^1 \Eb_{\rvx \sim p_t} [\|\boldsymbol{\phi}_t^\theta(\rvx) - \boldsymbol{\phi}_t(\rvx)\|^2 ] \dm t$} and discretization error due to step size $h$ in Theorem \ref{thm:combine}. The following assumptions are taken in the analysis.

\begin{algorithm}[t]
\caption{PGPS}\label{alg:train}
\begin{algorithmic}
\STATE {\bfseries Input:} Parameterized vector field $\boldsymbol{\phi}^\theta_t(\vx)$, Valid unnormalized path $\{\hat{p}_{t}\}_{t \in [0, 1]}$, Particles from initial distribution $\{\vx_{0, (i)}\}_{i=1\dots N}$, Maximum training steps $M$, Training threshold $\epsilon$, Learning rate $\eta$, Maximum time step $\Delta t'$, Particle step-size $\psi$;
\STATE Initialize $t \gets 0$;
\REPEAT
\FOR{$k = 1 \dots M$}
\STATE Gradient descent $\theta \gets \theta - \eta \nabla_\theta L_{\textbf{t}}(\theta)$;
\IF{$L_t(\theta) < \epsilon$}
\STATE \textbf{Break};
\ENDIF
\ENDFOR
\STATE $\Delta t \gets$ {\tt Adaptive Time Step}$[t, \boldsymbol{\phi}_t^\theta(\vx), \psi, \Delta t']$; \\ \texttt{~~// Algorithm \ref{alg:timestep} // }
\STATE Update $\{\vx_{t + \Delta t, (i)}\}\gets \{\vx_{t, (i)} + \Delta t \boldsymbol{\phi}_t^\theta(\vx_{t, (i)})\}$;
\STATE Update $t \gets t + \Delta t$;
\STATE (Optional) $\{\vx_{t, (i)}\}\gets$ \\{\tt Langevin Adjustment}$[\{\vx_{t, (i)}\}, \hat{p}_t]$;\\
\texttt{~~// Algorithm~\ref{alg:LD} //}
\UNTIL{t = 1}; 
\STATE {\bfseries Output:} Evolved particles $\{\vx_{1, (i)}\}_{i=1\dots N}$;
\end{algorithmic}
\end{algorithm}



\begin{assumption}[]\label{assumption}
~ \\\vspace{-.7cm}
\begin{enumerate}
\item[(1)] \textbf{Lipschitzness of $\boldsymbol{\phi}_t$ and  $\boldsymbol{\phi}_t^\theta$ on $\vx$ space}:
    There exists $K_1 < \infty$, such that $\| \boldsymbol{\phi}_t(\vx_1) - \boldsymbol{\phi}_t(\vx_2) \| \leq K_1 \|\vx_1 - \vx_2\|$ and $\| \boldsymbol{\phi}_t^\theta(\vx_1) - \boldsymbol{\phi}_t^\theta(\vx_2) \| \leq K_1 \|\vx_1 - \vx_2\|$ for any $\vx_1, \vx_2 \in \mathbb{R}^{d}$, $t\in [0, 1]$;
\item[(2)] \textbf{Lipschitzness of $\boldsymbol{\phi}_t^\theta$ on $t$ space}:
    There exists $K_2 < \infty$, such that $\|\boldsymbol{\phi}_{t_1}^\theta(\vx) - \boldsymbol{\phi}_{t_2}^\theta(\vx)\| \leq K_2 |t_2-t_1|$ for any $\vx \in \mathbb{R}^{d}$, $t_1, t_2\in [0, 1]$;
\item[(3)] \textbf{Finite vector field}:
There exists $K_3 < \infty$, such that $\|\boldsymbol{\phi}_t^\theta(\vx)\| \leq K_3$ for any $\vx \in \mathbb{R}^{d}$, $t\in [0, 1]$
\end{enumerate}
\end{assumption}

\begin{theorem}\label{thm:combine}
For two flows $\boldsymbol{\phi}_t^\theta(\rvx)$ and $\boldsymbol{\phi}_t( \rvx)$ under~\cref{assumption}, the Wasserstein distance between the distribution $p_{\hat{\rvx}_1^\theta}$ of PGPS generated samples according to dynamics \eqref{eq:descrete-PGPS} and the target distribution $p_{\rvx_1}$ is bounded as 
\begin{align}\footnotesize\label{eq:thm}
W_2(p_{\hat{\rvx}_1^\theta}, p_{\rvx_1})
&\leq \delta {  \sqrt{\exp(1 + 2 K_1)}} \notag \\
&\quad + \sqrt{h}\sqrt{\frac{C(\exp(1+K^2_1)-1)}{1+K_1^2}},
\end{align}
where $C = \frac{1}{2}K_2^2 + \frac{17}{2}K_1^2K_3^2 + 5K_1K_2K_3$.
\end{theorem}

There are two terms in the upper bound \cref{eq:thm} by the approximation error and the discretization error, respectively. The first term is related to the Lipschitzness assumption on $\boldsymbol{\phi}_t(\vx), \boldsymbol{\phi}^{\theta}_t(\vx)$ over $\vx$ space (Assumption \ref{assumption}(1)). It characterizes the error introduced due to the approximation of the vector field $\boldsymbol{\phi}_t$~(\cref{thm:continious_error}). The second term represents the error introduced by discretization, which is related to the Lipschitzness property with respect to $t$ and the finiteness of the vector field~(Assumption \ref{assumption}(2)-(3)). The proof of~\cref{thm:combine} can be found in~\cref{app:proof}.

\cref{thm:combine} indicates that with trained vector field of maximum error $\delta$ and discretized
with uniform step $h$ and the generated distribution is close to the target distribution with $W_2$-distance bounded by $\Oc(\delta) + \Oc(\sqrt{h})$. 
Therefore, we can improve the performance of the evolved particles by reducing the approximation error and/or refining the discretization.  {In the following, we illustrate that the training objective of minimizing loss function $L_t(\theta)$ in Equation \eqref{eq:pggfloss} is aligned with reducing the approximation error.}


{
  


Note that minimizing $L_t(\theta)$ is to solve the partial differential equation (PDE) in \eqref{eq:pggf}, which requires specifying the function space. Let $L^4(p_t)$ be the function space with norm $\|f\|_{L^4(p_t)} = (\int (f(\vx))^4 p_t(\vx) \dm \vx)^{1/4}$ and 
$W^{1, 4}(p_t) = \{f \in L^4(p_t): \frac{\partial}{\partial x_i} f(\vx) \in L^4(p_t) \}$ be a weighted Sobelov space. Denote by $\Psi_t = [W^{1, 4}(p_t)]^d$ a product space that contains the vector-valued functions of interest. Specifically, we made mild assumptions below
\begin{assumption}\label{assumption:pde}
($a$) $\boldsymbol{\phi}_t^\theta, \nabla \ln p_t \in \Psi_t$ for any $t \in [0, 1]$; and ($b$) $\sup_{t \in [0, 1]} \Eb_{\rvx \sim p_t}[\|\nabla \ln p_t(\rvx)\|^4 ] < \infty$. 
\end{assumption}
\begin{proposition}\label{pro:consistency} Under Assumption \ref{assumption:pde}, for any $\boldsymbol{\phi}^\theta_t$, there exists a vector-field $\boldsymbol{\phi}_t$ solution to PDE \eqref{eq:pggf} that 
\begin{align}
    \Eb_{\rvx \sim p_t} [\| \boldsymbol{\phi}_t^\theta(\rvx) - \boldsymbol{\phi}_t(\rvx) \|^2] \leq K L_t(\theta), 
\end{align}
where $K > 0$ is a universal constant factor and $L_t(\theta)$ is in \cref{eq:pggfloss} with infinite many particles following $p_t$. 
\end{proposition}
Proposition justifies the consistency of the proposed method, i.e., $\boldsymbol{\phi}_t$ can be well-approximated by minimizing the loss function \textit{under the infinite particles regime}. The impact of finite particles relies on the generalization analysis and is beyond the scope of the paper.

\section{Experiments} \label{sec:experiment}
We demonstrate the effectiveness of the proposed PGPS methods compared to LD, SVGD \cite{liu2016stein}, PFG \cite{dong2022particle} baselines. The number of iterations for each method is the same, where the Langevin Adjustment steps in PGPS are counted. The code to reproduce the experimetnal results can be found in our Github repository: \url{https://github.com/MingzhouFan97/PGPS}.

\begin{figure}[t]
\centering
\begin{subfigure}{0.49\linewidth}
    \includegraphics[width=1.05\textwidth]{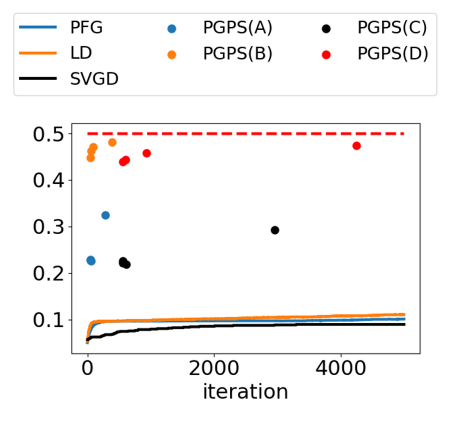}\vspace{-2mm}
    \caption{$\text{score}_1$}
    \label{fig:1d_lrr_ms}
\end{subfigure}
\begin{subfigure}{0.49\linewidth}
    \includegraphics[width=1.05\textwidth]{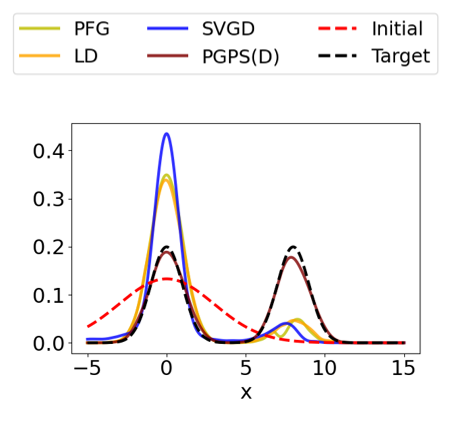}\vspace{-2mm}
    \caption{Mode seeking distribution}
    \label{fig:1d_prob_ms}
\end{subfigure}
\begin{subfigure}{0.49\linewidth}
    \includegraphics[width=1.05\textwidth]{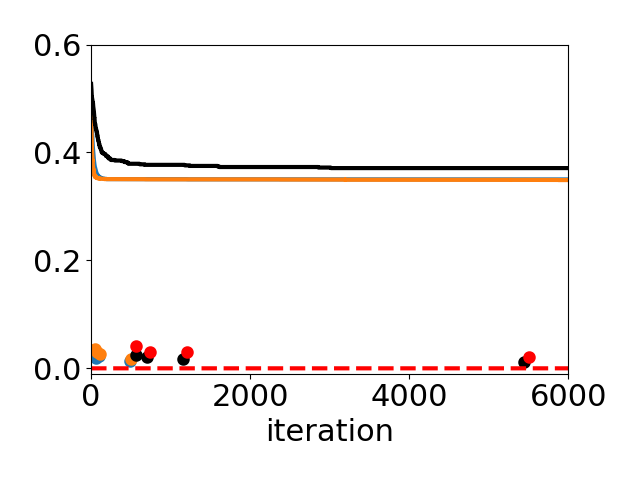}\vspace{-2mm}
    \caption{$\text{score}_2$}
    \label{fig:1d_lrr_wr}
\end{subfigure}
\begin{subfigure}{0.49\linewidth}
    \includegraphics[width=1.05\textwidth]{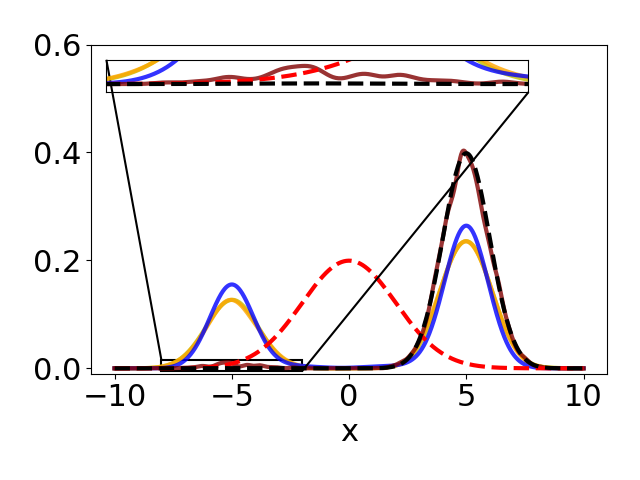}\vspace{-2mm}
    \caption{Sensitivity distribution}
    \label{fig:1d_prob_wr}
\end{subfigure}\vspace{-3mm}
\caption{The performances of different methods: (a, c) $\text{score}_1$ and $\text{score}_2$ indicating the mode capture ability with the true score illustrated by the red dashed line; (b, d) KDE estimated probability distributions for different methods. The letter following PGPS indicates different hyperparameters. (A): $\alpha=0$, $\beta = 1$, $\text{steps} = 0$ (B): $\alpha=1$, $\beta = 0.8$, $\text{steps} = 0$ (C): $\alpha=0$, $\beta = 1$, $\text{steps} = 10$ (D): $\alpha=1$, $\beta = 0.8$, $\text{steps} = 10$, where `$\text{steps}$' indicates the number of performed Langevin Adjustment steps. We report the performance of PGPS with $\psi\in\{0.5, 0.1, 0.05, 0.01\}$.}\label{fig:1d}\vspace{-3mm}
\end{figure}

\subsection{Gaussian Mixture Target Distribution}\label{sec:Gaussian}

We study the mode-seeking and weight-estimation capabilities of the proposed PGPS for Gaussian mixture target distributions, compared to LD, SVGD, and PFG gradient-flow particle-based benchmarks. 

\subsubsection{Mode Discovery Missing}
Given initial distribution $\mathcal{N}(0, 3^2)$ and target distribution of a mixture of two Gaussian distributions $\mathcal{N}(0, 1)$ and $\mathcal{N}(8, 1)$ with equal weights, we investigate whether the methods can effectively discover both modes. 

Note that the left mode $\Nc(0, 1)$ of the target mixture is automatically discovered by the initial distribution $\Nc(0, 3^2)$. Define $\text{score}_1 = \frac{\sum_{i=1}^N \mathbb{I}(x_{t,(i)} > 5)}{N}$ to capture the rates of the samples discovering the right mode $\Nc(8, 1)$ by moving across threshold 5. The $\text{score}_1$ is shown in Figure~\ref{fig:1d_lrr_ms}, where the dashed true score is $\Pb_{\text{target}}(\rvx > 0.5) \approx 0.499$. 
Note that the performances of PGPS methods are scattered because the method may require different adaptive iterations for different hyperparameter choices. With the fact that the intermediate state of the PGPS particle is not meaningful, scatter plots are selected rather than lines for LD, SVGD, and PFG.
As shown in Figure~\ref{fig:1d_lrr_ms}, PGPS recovers the right mode faster and better with $\text{score}_1$ close to the true score $0.49$, yet the benchmarks fail. Figure~\ref{fig:1d_prob_ms} corroborates the finding by visualizing the output distribution of the sample methods, where PGPS-generated distribution is closer to the target. 

\subsubsection{False Mode Discovery -- Sensitivity}
The benchmarks not only fail to effectively discover modes but are also sensitive to the target distribution and may lead to false discovery, i.e., they may focus on some negligible mode. 

Given initial distribution $\mathcal{N}(0, 2^2)$ and target distribution of a mixture of two Gaussian distributions $\mathcal{N}(-5, 1)$ and $\mathcal{N}(5, 1)$, where the left mode has an extremely small weight $0.001$ and the right mode has weight $0.999$. As shown in Figure~\ref{fig:1d_prob_wr}, the left mode is negligible and the target distribution is visually indistinguishable from a Gaussian distribution. 

Define $\text{score}_2 = \frac{\sum_{i=1}^N \mathbb{I}(x_{t,(i)} < 0)}{N}$ to capture the rates of the samples focusing on the negligible left mode $\Nc(-5, 1)$. The $\text{score}_2$ is shown in Figure~\ref{fig:1d_lrr_wr}, where the dashed true score is $\Pb_{\text{target}}(\rvx < 0) \approx 0.001$. We observe that the benchmarks have a relatively large $\text{score}_2$, which indicates they are very sensitive w.r.t. the target distribution. A negligible perturbation from the Gaussian target may lead to these methods focusing on a negligible mode. In contrast, the proposed PGPS is less sensitive with $\text{score}_2$ close to the desired value $0.001$. Figure~\ref{fig:1d_prob_wr} corroborates the finding by visualizing the output distribution of the methods. 

Compared to the gradient-flow-based benchmarks solely relying on the target distribution and its gradient, the proposed PGPS method follows a smooth LwS path instead, and is indeed less sensitive with better sampling quality.

\begin{table*}[]
\centering
\caption{Average Expected Calibration Error~(ECE) and Accuracy~(ACC) on UCI datasets over five independent runs}
\label{tab:UCI}\vspace{-3mm}
\scalebox{0.7}{\begin{tabular}{c|cccc|cccc}
\hline
 \multicolumn{1}{c}{} & \multicolumn{4}{c}{Expected Calibration Error (ECE) $\downarrow$} & \multicolumn{4}{c}{Accuracy (ACC) $\uparrow$} \\
 \hline 
 & \textbf{PGPS}     &  SVGD &    SGLD&       PFG&      \textbf{PGPS}   &  SVGD &    SGLD&       PFG\\
 \hline
SONAR &$0.2517 \pm0.057$&${0.1712} \pm0.020$&$0.3394 \pm0.049$&$\textbf{0.1678} \pm0.050$&$\textbf{0.7981} \pm0.023$&$0.7962 \pm0.016$&$0.7942 \pm0.024$&$0.7673 \pm0.033$\\
WINEWHITE&$\textbf{0.0750} \pm0.011$&$0.0988\pm0.012$&${0.0935}\pm0.024$&$0.0876\pm0.018$&$0.4520\pm0.010$&$0.4520\pm0.010$&$\textbf{0.4831}\pm0.049$&$0.4520\pm0.010$\\
WINERED&$\textbf{0.0366}\pm0.005$&$0.0402\pm0.004$&$0.0868\pm0.029$&${0.0449}\pm0.005$&$\textbf{0.5938}\pm0.018$&$0.5770\pm0.018$&$0.5107\pm0.096$&$0.5723\pm0.019$\\
AUSTRALIAN&${0.1703} \pm 0.066$&${0.1713} \pm0.064$&$0.3517 \pm0.078$&$\textbf{0.1457} \pm0.047$&$0.8620 \pm0.009$&$0.8626 \pm0.006$&$0.7362 \pm0.157$&$\textbf{0.8643} \pm0.006$\\
HEART&$\textbf{0.4579} \pm0.071$&$0.5117 \pm0.064$&$0.5110 \pm0.114$&$0.4887 \pm0.089$&$\textbf{0.2556} \pm0.142$&$0.1801 \pm0.042$&$0.2384 \pm0.135$&$0.1762 \pm0.033$\\
GLASS&$\textbf{0.1142} \pm0.008$&${0.1155}\pm0.006$&$0.2157\pm0.025$&$0.1289\pm0.021$&$\textbf{0.5850}\pm0.080$&$0.5383\pm0.076$&$0.4561\pm0.152$&$0.4505\pm0.071$\\
COVERTYPE&$\textbf{0.0743} \pm 0.016$&${0.0950} \pm0.012$&$0.1301 \pm0.038$&$0.0926 \pm0.078$&$\textbf{0.5899} \pm0.095$&$0.4867 \pm0.006$&$0.5221 \pm0.084$&$0.5088 \pm0.053$\\
 \hline
\end{tabular}}
\end{table*}







\subsubsection{Weight Recovery}

\begin{figure}[b!]
    \vspace{-0mm}
    \includegraphics[width=0.9\linewidth]{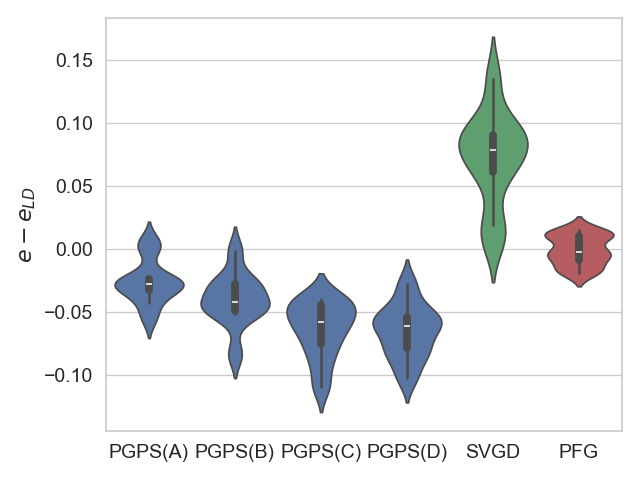}\vspace{-3mm}
    \caption{The weight mismatch error. 
    The letter after PGPS indicates different hyperparameters. (A): $\alpha=0$, $\beta = 1$, $\text{steps} = 0$ (B): $\alpha=0$, $\beta = 0.5$, $\text{steps} = 0$ (C): $\alpha=0$, $\beta = 1$, $\text{steps} = 100$ (D): $\alpha=0$, $\beta = 0.5$, $\text{steps} = 100$, where `$\text{steps}$' is the number the Langevin Adjustment steps.} 
    \label{fig:highd_diff}
\end{figure}

We investigate the capability of the proposed PGPS method in estimating the corresponding weights besides detecting modes. The target distribution is a mixture of four 8-dimensional isometric Gaussian distributions $\{\mathcal{N}(\vmu_j, 0.15^2 \mathbf{I}_8)\}$ and randomly generated weights; and the initial distribution is $\mathcal{N}(\mathbf{0}, \mathbf{I}_8)$.


For generated samples $\{\vx_i\}_{i = 1}^N$, 
define the estimated weight $\hat{\omega}_j := \frac{\sum_{i=1}^N\mathbb{I}(\|\vx_i-\vmu_j\| < 1)}{N}$. We evaluate the weight mismatch by $e := \sqrt{\sum_{j=1}^4 (\hat{\omega}_j - \omega_j)^2}$, where $\omega_j := \Pb_{\text{target}}(\|\rvx - \vmu_j\|) < 1)$ is the ground truth. Smaller error $e$ indicates more accurate weight estimation. 

We take LD (a realization of the Wasserstein gradient flow) as a baseline, and denote its weight mismatch error by $e_{LD}$. In Figure~\ref{fig:highd_diff}, we demonstrate the distribution of the difference between the weight mismatch error $e$ of a method and the baseline $e_{LD}$ averaged over $10$ independent experiments. 
 {While the baseline $e_{LD}$ has an average value of 0.3314, t}he proposed PGPS methods consistently outperform LD with the distributions of $e - e_{LD}$ being significantly less than $0$. The performance of PFG is similar to LD because of the same Wasserstein gradient flow nature, while SVGD performs worse than LD for this task. The inferior performance of SVGD is mainly due to the curse of dimensionality, which makes it difficult for the particles to escape from the trapping modes~\cite{liu2022grassmann}.

\subsection{Bayesian Neural Network Inference}
We further test PGPS methods for the Bayesian Neural Network~(BNN) inference tasks. BNNs, which model the parameters of NNs as random variables to derive predictive posteriors for prediction, are usually considered to be difficult inference targets because of their non-concave likelihoods~\cite{li2018visualizing}. 
The proposed PGPS methods, with a stronger ability to discover the modes and recover their weights, achieve better inference performance.

\subsubsection{UCI dataset}\label{sec:UCI}
We conduct BNN inference for UCI datasets~\cite{Dua:2019}, where the neural network (NN) has one hidden layer with 32 hidden neurons and \emph{Sigmoid} activation. More details of the experimental setup can be found in the Appendix \ref{app:bnn}. 



We report the averaged testing Expected Calibration Error~(ECE) and testing accuracy~(ACC) in~\cref{tab:UCI}, where ECE represents the calibration ability of the uncertain prediction by comparing the difference in prediction accuracy and prediction uncertainty for the test samples. 
The proposed PGPS methods achieve the best performance across most of the benchmark datasets with lower ECE and higher ACC, compared with SVGD, SGLD, and PFG baselines. 


\subsubsection{Noisy MNIST dataset}\label{sec:mnist}
Robustness is another desired property of learning Bayesian models. It is expected that Bayesian models would give more reasonable predictions with uncertainty quantification~(UQ) when facing out-of-distribution data. We benchmark the prediction and UQ performance of the proposed PGPS methods for learning BNNs 
on the MNIST dataset~\cite{deng2012mnist}. 

To test the robustness of inferred models, we create perturbation by injecting additive Gaussian noise into the test MNIST images. Ensembles of 10 learned BNNs (i.e., 10 particles) are considered for evaluating competing inference methods. 
The performances are evaluated by negative log-likelihood (NLL), ACC, and ECE in~\cref{tab:mnist}. We can observe that the proposed PGPS method is again the best-performing inference method on all the metrics with the perturbed test data. SGLD is slightly better in NLL by $0.02$ but with a large standard deviation of $0.127$. 

\begin{table}[]
\centering
\caption{Average negative log-likelihood~(NLL), ACC, and ECE on Noisy MNIST data over five independent runs}
\label{tab:mnist}\vspace{-3mm}
\scalebox{0.7}{\begin{tabular}{c|cccc}
\hline 
& \textbf{PGPS} & SVGD & SGLD & PFG \\
\hline
NLL $\downarrow$ & $1.8202 \pm 0.019$ & $1.8285 \pm 0.040$ & $\textbf{1.8184} \pm 0.127$ & $2.0171 \pm 0.014$ \\
ACC $\uparrow$ & $\textbf{0.8788} \pm 0.017$ & $0.8282 \pm 0.047$ & $0.6419 \pm 0.130$ & $0.7119 \pm 0.027$ \\
ECE $\downarrow$ & $\textbf{0.1716} \pm 0.012$ & $0.1941 \pm 0.020$ & $0.2183 \pm 0.030$ & $0.1752 \pm 0.003$ \\
\hline
\end{tabular}}
\vspace{-0.3cm}
\end{table}

\subsubsection{Training-free PSPG}\label{app:ALD}



We compare the standard PGPS and the training-free PGPS as discussed in Section~\ref{sec:learning-vector-field} using the same \emph{Log-weighted Shrinkage Path} on the noisy MNIST data as in~\cref{sec:mnist}. The performance of standard PGPS and training-free PGPS (tf-PGPS) is reported in Table~\ref{tab:ab}). We can observe that standard PGPS achieves better performance among almost all metrics and the number of particles than tf-PGPS. For the cases where tf-PGPS is better, their performances are almost indistinguishable.
Interestingly, NLL increases as the number of particles goes up. We reason this by the fact that when using more particles for estimation, the predictions tend to fit the target posterior distribution better and lead to higher ACC but higher NLL as well.
\begin{table}[]
\centering
\caption{Averaged NLL, ACC, and ECE on Noisy MNIST data over five independent runs}
\label{tab:ab}\vspace{-3mm}

\scalebox{0.75}{
\begin{tabular}{c c|cc}
\hline 
\# particles & & PGPS & tf-PGPS \\
\hline
\multirow{3}{*}{10} & NLL $\downarrow$ & $\textbf{1.8171} \pm0.0168$&$1.8238 \pm0.0251$ \\
&ACC $\uparrow$ &$\textbf{0.8683} \pm0.0193$&$0.8380 \pm0.0504$\\
&ECE $\downarrow$ &$0.1680 \pm0.0132$&$\textbf{0.1659} \pm0.0083$\\
\hline
\multirow{3}{*}{50} & NLL&$1.8473 \pm0.0101$&$\textbf{1.8467} \pm0.0108$ \\
&ACC&$\textbf{0.9006} \pm0.0071$&$0.8672 \pm0.0298$\\
&ECE&$\textbf{0.1807} \pm0.0034$&$0.1890 \pm0.0071$\\
\hline
\multirow{3}{*}{100} & NLL&$\textbf{1.8763} \pm0.0087$&$1.9010 \pm0.0135$ \\
&ACC&$\textbf{0.9182} \pm0.0036$&$0.8956 \pm0.0259$\\
&ECE&$\textbf{0.1959} \pm0.0024$&$0.1986 \pm0.0068$\\
\hline
\end{tabular}
}
\vspace{-0.5cm}
\end{table}

\section{Conclusion}
In this paper, we proposed a novel path-guided particle-based sampling (PGPS) method and a \emph{Log-weighted Shrinkage} path as a partition-function-free path that guides the particles moving from an initial distribution to the target distribution. We theoretically analyzed the performance of PGPS under the Wasserstein distance and characterized the impact of approximation error and discretization error on the quality of the generated samples. We conduct extensive experiments to test the PGPS methods in seeking the modes of the target distribution in sampling tasks, and the inference performance in terms of testing accuracy and calibration/uncertainty quantification in Bayesian learning tasks. The proposed PGPS methods perform consistently and considerably better than LD, SVGD, and PFG benchmarks in the experiments. 

A limitation of the standard PGPS method is the requirement of training neural networks, similar to the PFG and other learning-required benchmarks. We propose training-free PGPS as an immediate solution, which is slightly worse than the training-based PGPS but more efficient. 

A better density path design in PGPS that leverages the structure of the target distribution and analysis of training-free PGPS of its convergence to the target distribution are interesting future directions with both theoretical and practical importance. 



\newpage

\section*{Acknowledgements}
R. Zhou would like to thank Chunyang Liao for the helpful discussion on Proposition 4.4. This work was supported in part by the U.S. National Science Foundation~(NSF) grants SHF-2215573, IIS-2212419, and DMS-2312173; and by the U.S. Department of Engergy~(DOE) Office of Science, Advanced Scientific Computing Research (ASCR) M2DT Mathematical Multifaceted Integrated Capability Center~(MMICC) under Award B\&R\# KJ0401010/FWP\# CC130, program manager W. Spotz. Portions of this research were conducted with the advanced computing resources provided by Texas A\&M High Performance Research Computing.

\section*{Impact Statement}
This paper presents a new Bayesian inference algorithm for efficient Bayesian learning, which would be critical in small-data scenarios that require uncertainty quantification. It can motivate advances in science, engineering, and biomedicine, for example in bioinformatics, materials science, and high-energy physics.








\bibliography{reference}
\bibliographystyle{icml2024}

\newpage
\onecolumn
\appendix
\section{Related Works}\label{app:related}
Wasserstein gradient flow aims at building gradient flow where the density follows the steepest descent path of some objective functional of density function under the Wasserstein metric. Sampling is fulfilled once the descent path converges to the target distribution, which is the minimizer of the objective function. A popular objective function of density is the KL-divergence between the density and the target distribution, and the flow is thus called KL Wasserstein gradient flow. While it is intractable in general to derive the KL Wasserstein gradient flow, i.e., the flow does not have a closed-form, \citet{wang2022projected} resorted to Kernel Density Estimation (KDE) to estimate the gradient flow and uses Euler discretization to update the samples. However, it suffers from the curse of dimensionality, i.e. the kernel matrix would tend to be diagonal as dimensionality increases, due to the nature of kernels, which leads to inaccurate density estimation. 

Aside from Euler discretization, the Jordan, Kinderlehrer, and Otto~(JKO) scheme, aiming at finding a JKO operator that minimizes the target functional as well as the movement of the particles in each step, has been broadly applied to discretize the Wasserstein gradient flow. \citet{alvarez2021optimizing} and~\citet{mokrov2021large} applied a series of Input Convex Neural Networks (ICNN, \citet{amos2017input}) to model the gradient flow to ensure the convexity of the potential function in JKO scheme.

While the popularity of Stein Variational Gradient Descent~(SVGD)~\citep{liu2016stein} arises as a particle-based VI method, it can also be viewed as a specific type of gradient flow w.r.t. KL-divergence by determining the gradient $\boldsymbol{\phi}_t(\rvx)$ that is the steepest descent direction under the kernelized Stein's Discrepancy~\citep{chwialkowski2016kernel} by the reproducing kernel Hilbert space (RKHS)~\citep{liu2017stein}. However, the curse of dimensionality for the kernel-based methods leads to the particle collapse in SVGD~\citep{ba2021understanding}, i.e., variance collapse. Currently, there are two major methods to tackle the curse of dimensionality. Projecting the inference space to a lower dimension can naturally avoid high-dimensional VI. While \citet{chen2020projected} projected the dynamics into a lower dimensional subspace and theoretically proved the asymptotically converging performance of the projected SVGD, \citet{gong2020sliced} proposed sliced kernel Stein discrepancy that projects the particle dynamics into a single dimensional subspace. More recently, \citet{liu2022grassmann} proposed Grassmann SVGD that also considers a low-dimensional projection and is claimed to be more efficient than projected SVGD~\citep{chen2020projected} without the need for costly eigenvector decomposition. Another type of popular method leverages the Universal Approximation Theorem of Neural Networks (NNs)~\citep{hornik1989multilayer} and defines more general discrepancy.  \citet{di2021neural} proposed to minimize Stein's discrepancy based on the NNs, instead of functions drawn from RKHS like SVGD. \citet{grathwohl2020learning} proposed to learn a single energy function based on Stein's Discrepancy for energy-based models, while \citet{hu2018stein} tried to learn a transport plan based on Stein's Discrepancy or more general f-divergence. \citet{dong2022particle} modified the regularization term of the loss function to a preconditioned version but it needs to calculate the Jacobian of the target density and in turn time-consuming. 

\section{Implementation of LwS Path}\label{app:gradient}
Though it is possible to fully depend on the AutoGrad functionality of the machine learning packages, a relatively closed form of the gradient and derivatives of our \emph{Log-Shrinkage Path}, $\ln \hat{p}_t^{\text{LwS}}(\vx) =({1-t}) \ln p_0\left((1-\alpha t)\vx\right)+ t \ln \hat{p}_1 \left(\frac{\vx}{\beta +(1 - \beta) t}\right)$, would lead to better calculation quality and faster computational speed.

Denote $\vx_a = (1-\alpha t)\vx$, $\vx_b = \frac{\vx}{\beta + (1-\beta)t}$. The gradient of our path $\hat{p}_t^{\text{LwS}}(\vx)$ at time $t$ would be
\begin{equation}
\nabla \ln \hat{p}_t^{\text{LwS}}(\vx) = (1-t)(1-\alpha t)\nabla \ln p_0(\vx_a) + \frac{t}{\beta + (1-\beta) t}\nabla \ln \hat{p}_1(\vx_b),
\end{equation}
and the derivative would be
\begin{equation}
\frac{\diff}{\diff t} \ln \hat{p}_t^{\text{LwS}}(\vx) = - \ln p_0(\vx_a) + \ln \hat{p}_1(\vx_b) - \alpha  (1-t)\vx \cdot \nabla \ln p_0(\vx_a) -\frac{(1-\beta)t \vx \cdot \nabla \ln \hat{p}_1(\vx_b)}{(\beta + (1-\beta)t)^2},
\end{equation}
where $(\cdot)$ denotes inner product. 

In our training target of~\cref{eq:pggfloss}, one critical part is the divergence $\nabla \cdot \boldsymbol{\phi}^\theta$ of the approximated vector field $\boldsymbol{\phi}^\theta$. While \cite{dong2022particle} proposed to use an efficient computational estimation derived by the integration-by-parts technique, a close form of divergence can be derived for relatively simple NN implementation. Specifically, for the one hidden layer, $H$ hidden neuron, $D$ dimensional input, \emph{sigmoid} activation MLP we used in this work, $\boldsymbol{\phi}^\theta = W_2\sigma(W_1 \vx + b_1) + b_2$, the close form of divergence would be
\begin{equation}
\nabla_\vx \cdot \boldsymbol{\phi}^\theta(\vx) = \sum_{i=1}^D (\nabla_\vx \boldsymbol{\phi}_{(i)}^\theta(\vx))_i = trace(W_2 \text{diag}(\vx_g) W_1) = \sum W_2 \otimes W_1^T \otimes (\boldsymbol{1}_D \vx_g^T),
\end{equation}
where $\sigma$ is the \emph{sigmoid} function, $\vx_h = \sigma(W_1 \vx + b_1)$ is the output of the first layer, $\otimes$ denotes entry-wise product, $\vx_g =  \vx_h \otimes (\boldsymbol{1}_H - \vx_h)$, $\boldsymbol{1}_D$ and $\boldsymbol{1}_H$ denotes all one matrix with size of $D \times 1$ and $H \times 1$, respectively, and the ``$\sum$'' sign in the last equation denotes summation along both dimensions.
{  
\section{Influence of Hyperparameter in LwS Path}\label{app:hyper}

To show the impact of hyperparameter choices, we here give an example with the same target as the motivating example in Figure 1(a), a mixture of two Gaussian distributions with equal weights. We apply the training-free version of PGPS discretized with a constant time step of $0.01$, i.e. $99$ intermediate distributions, and $30$ LD steps performed for each intermediate distribution to ensure that different setups consume the same computational resource. 

With the results illustrated in Figure~\ref{fig:hyper}, it can be observed that the hyperparameter choices can influence the sample quality with the same computational demand. The no-shrinkage setup leads to the worst performance and the hyperparameter choices that incorporate shrinkage capture the mode to the right much better.

\begin{figure}[t]
\centering
\begin{subfigure}{0.32\linewidth}
    \includegraphics[width=\textwidth]{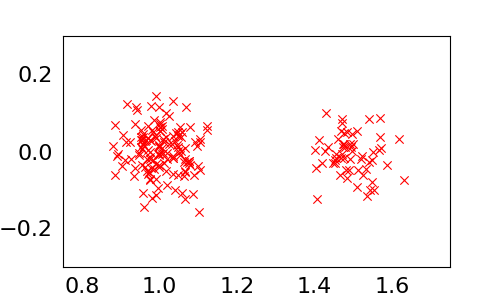}
    \caption{$\alpha=0, \beta=0.1$}
\end{subfigure}
\begin{subfigure}{0.32\linewidth}
    \includegraphics[width=\textwidth]{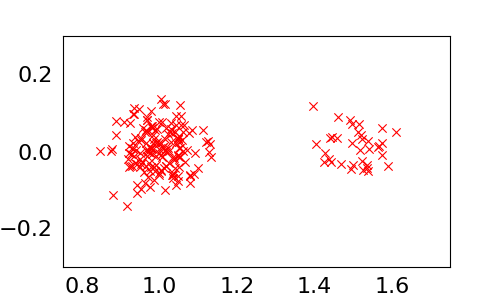}
    \caption{$\alpha=0, \beta=0.5$}
\end{subfigure}
\begin{subfigure}{0.32\linewidth}
    \includegraphics[width=\textwidth]{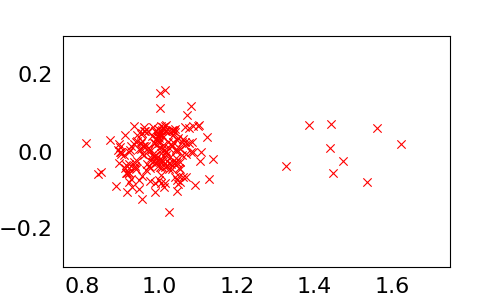}
    \caption{$\alpha=0, \beta=1$}
\end{subfigure}
\begin{subfigure}{0.32\linewidth}
    \includegraphics[width=\textwidth]{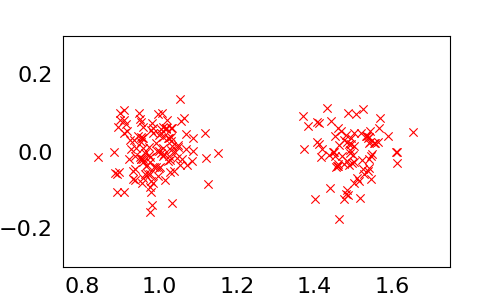}
    \caption{$\alpha=0.5, \beta=0.1$}
\end{subfigure}
\begin{subfigure}{0.32\linewidth}
    \includegraphics[width=\textwidth]{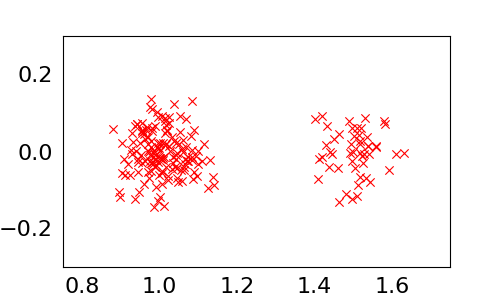}
    \caption{$\alpha=0.5, \beta=0.5$}
\end{subfigure}
\begin{subfigure}{0.32\linewidth}
    \includegraphics[width=\textwidth]{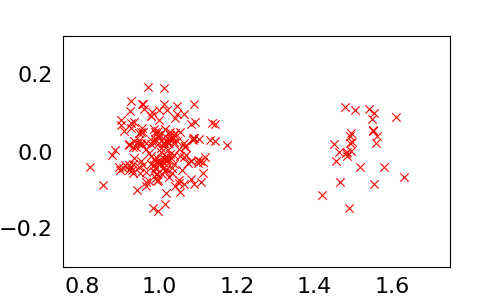}
    \caption{$\alpha=0.5, \beta=1$}
\end{subfigure}
\begin{subfigure}{0.32\linewidth}
    \includegraphics[width=\textwidth]{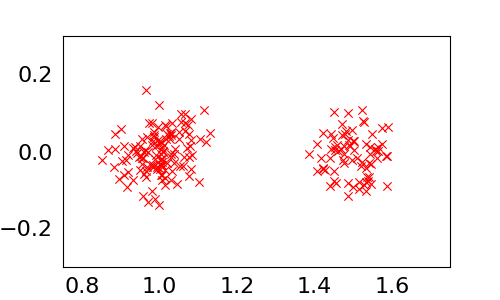}
    \caption{$\alpha=1, \beta=0.1$}
\end{subfigure}
\begin{subfigure}{0.32\linewidth}
    \includegraphics[width=\textwidth]{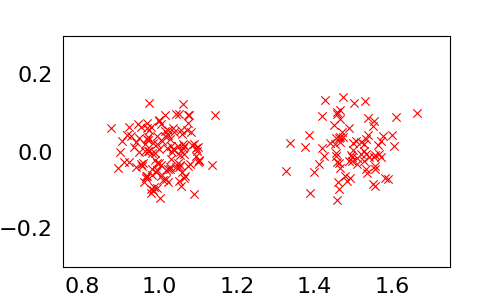}
    \caption{$\alpha=1, \beta=0.5$}
\end{subfigure}
\begin{subfigure}{0.32\linewidth}
    \includegraphics[width=\textwidth]{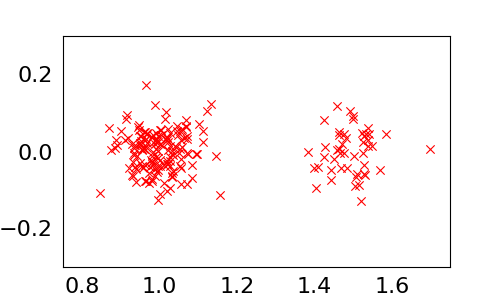}
    \caption{$\alpha=1, \beta=1$}
\end{subfigure}
\caption{Particles for tf-PGPS following the LwS-path with different hyperparameters discretized with a constant time step of $0.01$, $30$ LD steps for each intermediate distribution. With the same computational demand, the hyperparameter choices can influence the sample quality. The no-shrinkage setup $(\alpha=0, \beta=1)$ leads to the worst performance and the hyperparameter choices that incorporate shrinkage capture much better the mode on the right.}\label{fig:hyper}
\end{figure}

}

\section{Proofs}\label{app:proof}

\subsection{Proof of~\cref{prop:equality}}

\begin{proposition}
    For a given partition-free density path $\{\hat{p}_t\}$, the gradient flow guided by the vector field $\boldsymbol{\phi}_t(\rvx)$ following the Fokker-Planck equation~\eqref{eq:continuity} satisfies:
    \begin{equation}
    r(\vx, \boldsymbol{\phi}_t) - \mathbb{E}_{\rvx \sim p_t}\left[\frac{\partial \ln\hat{p}_t(\rvx)}{\partial t}\right]=0,
    \end{equation}
    where $r(\vx, \boldsymbol{\phi}_t) = \frac{\partial \ln\hat{p}_t(\vx)}{\partial t}  + (\nabla \ln{\hat{p}_t}(\vx) + \nabla) \cdot \boldsymbol{\phi}_t(\vx)$.
\end{proposition}

\begin{proof}
We start with the Fokker-Planck equation:
\begin{equation}\label{eq:app_contiu}
    \frac{\partial}{\partial t} p_t(\vx) = - \nabla \cdot (p_t(\vx) \boldsymbol{\phi}_t(\vx)),
\end{equation}
The right-hand side~(RHS) of~\eqref{eq:app_contiu} can be derived to be:
\begin{equation}\label{eq:RHS}
    -\nabla \cdot (p_t(\vx) \boldsymbol{\phi}_t(\vx))
    = - p_t(\vx) [\nabla \ln{p_t}(\vx) \cdot \boldsymbol{\phi}_t(\vx) + \nabla \cdot \boldsymbol{\phi}_t(\vx)].
\end{equation}
Though it is usually non-trivial to find the derivative of the path with respect to time, it is clear that $\nabla \ln{p_t} = \nabla \ln{\hat{p}_t} + \nabla \ln{\int \hat{p}_t \diff \vx} = \nabla \ln{\hat{p}_t}$. Equation~\eqref{eq:RHS} can then be further transformed into:
\begin{equation}\label{eq:RHSM}
    -\nabla \cdot (p_t(\vx) \boldsymbol{\phi}_t(\vx))
    = - p_t(\vx) [\nabla \ln{\hat{p}_t}(\vx) \cdot \boldsymbol{\phi}_t(\vx) + \nabla \cdot \boldsymbol{\phi}_t(\vx)].
\end{equation}

On the other hand, the left-hand side~(LHS) of~\eqref{eq:app_contiu} can be derived as:
\begin{equation}
\begin{aligned}\label{eq:LHS}
    &\frac{\partial}{\partial t} p_t(\vx) = \frac{\partial}{\partial t} \frac{\hat{p}_t(\vx)}{\int \hat{p}_t(\vx) \diff \vx}\\
    = &\frac{\int \hat{p}_t(\vx) \diff \vx \frac{\partial}{\partial t}\hat{p}_t(\vx) - \hat{p}_t(\vx)\frac{\partial}{\partial t}\int \hat{p}_t(\vx) \diff \vx}{(\int \hat{p}_t(\vx) \diff \vx)^2}\\
    = &\frac{\frac{\partial}{\partial t}\hat{p}_t(\vx)}{\int \hat{p}_t(\vx) \diff \vx} - \frac{\hat{p}_t(\vx)}{\int \hat{p}_t(\vx) \diff \vx}\int \frac{\frac{\partial}{\partial t} \hat{p}_t(\vx)}{\int \hat{p}_t(\vx) \diff \vx}\diff \vx\\
    = &\frac{\hat{p}_t(\vx)}{\int \hat{p}_t(\vx) \diff \vx}\frac{\partial}{\partial t} \ln \hat{p}_t(\vx) - p_t(\vx)\int \frac{\hat{p}_t(\vx)}{\int \hat{p}_t(\vx) \diff \vx} \frac{\partial}{\partial t} \ln\hat{p}_t(\vx) \diff \vx\\
    = &p_t(\vx)(\frac{\partial \ln\hat{p}_t(\vx)}{\partial t} - \int p_t(\vx)\textbf{} \frac{\partial \ln\hat{p}_t(\vx)}{\partial t} \diff \vx)\\
    = &p_t(\vx) \left(\frac{\partial \ln\hat{p}_t(\vx)}{\partial t} - \mathbb{E}_{\rvx \sim p_t}\left[\frac{\partial \ln\hat{p}_t(\rvx)}{\partial t}\right] \right).
\end{aligned}
\end{equation}
Substituting equations~\eqref{eq:RHSM} and~\eqref{eq:LHS} to~\cref{eq:app_contiu} and $p_t(\vx)$ on both sides and we have the desired result
\begin{equation}
 \nabla \ln{\hat{p}_t}(\vx) \cdot \boldsymbol{\phi}_t(\vx) + \nabla \cdot \boldsymbol{\phi}_t(\vx) = \frac{\partial \ln\hat{p}_t(\vx)}{\partial t} - \mathbb{E}_{\rvx \sim p_t}\left[\frac{\partial \ln\hat{p}_t(\rvx)}{\partial t}\right]. 
    \end{equation}
\end{proof}

\subsection{Proof of~\cref{thm:combine}}
Before showing the quality analysis of the evolved particles with the discretized algorithm, we first evaluate the impact of error between the numerical approximation $\boldsymbol{\phi}^\theta_t(\vx)$ and the true $\boldsymbol{\phi}_t(\vx)$ satisfying~\cref{eq:pggf}.

\begin{lemma}[Proposition~3 of \citet{albergo2023building}] \label{thm:continious_error}
For two flows $\boldsymbol{\phi}_t^\theta(\vx)$ and $\boldsymbol{\phi}_t(\vx)$ under Assumption~\ref{assumption}, the Wasserstein distance between the distribution $p_{\rvx_1^\theta}$ of random variable $\rvx_1^\theta = \rvx_0 + \int_0^1 \boldsymbol{\phi}_t^\theta(\rvx_t) \diff t $, and the distribution $p_{\rvx_1}$ of $\rvx_1 = \rvx_0 + \int_0^1 \boldsymbol{\phi}_t(\rvx_t) \diff t $ is bounded:
\begin{equation}
W^2(p_{\rvx_1^\theta}, p_{\rvx_1}) \leq \delta^2 \exp(1 + 2K_1).
\end{equation}
\end{lemma}

Due to the necessary discretization involved in particle evolution, another error factor arises. We here give the analysis of the discretization error.

\begin{lemma}\label{thm:discretization_error}
For a trained flow $\boldsymbol{\phi}_t^\theta(\vx)$ under~\cref{assumption}, the Wasserstein distance between the distribution $p_{\rvx_1}$ of random variable $\rvx_1 = \rvx_0 + \int_0^1 \boldsymbol{\phi}_t^\theta(\rvx_t) \diff t$ and the distribution $p_{\hat{\rvx}_1}$ of random variable $\hat{\rvx}_1$ generated by constant discretization $\hat{\rvx}_{(n+1)h} = \hat{\rvx}_{nh} + \boldsymbol{\phi}_{nh}^\theta(\hat{\rvx}_{nh})$ with step-size $h$ is bounded:
\begin{equation}
W^2(p_{\rvx_1}, p_{\hat{\rvx}_1})\leq h\frac{C(\exp(1+K^2_1)-1)}{1+K_1^2},
\end{equation}
where $C = \frac{1}{2}K_2^2 + \frac{17}{2}K_1^2K_3^2 + 5K_1K_2K_3$ is a constant.
\end{lemma}
\begin{proof}
Here we consider the discretization error
\begin{align}
&W^2(p_{\rvx_{t+h}}, p_{\hat{\rvx}_{t+h}})\\
\leq&\mathbb{E}_{\gamma}\|\rvx_{t+h} - \hat{\rvx}_{t+h}\|^2\\
=&\mathbb{E}_{\gamma}\|\rvx_{t} + (\rvx_{t+h} - \rvx_t) - [\hat{\rvx}_t + h \boldsymbol{\phi}_t( \hat{\rvx}_t)]\|^2\\
\leq&\mathbb{E}_{\gamma}\|[\rvx_{t} - \hat{\rvx}_t] + [\rvx_{t+h} - \rvx_t - h \boldsymbol{\phi}_t( \hat{\rvx}_t)]\|^2\\
\leq&(1+\lambda)\mathbb{E}_{\gamma}\|\rvx_{t} - \hat{\rvx}_t\|^2 + (1+\frac{1}{\lambda}) \mathbb{E}_{\gamma}\|\int_t^{t+h}\boldsymbol{\phi}(t', \rvx_{t'})dt' - h\boldsymbol{\phi}_t( \hat{\rvx}_t)\|^2\label{eq:dis_sec_1}
\end{align}
Now we bound the second term:
\begin{align}
&\mathbb{E}_{\gamma}\|\int_t^{t+h}\boldsymbol{\phi}(t', \rvx_{t'})dt' - h\boldsymbol{\phi}_t( \hat{\rvx}_t)\|^2\\
\leq&\mathbb{E}_{\gamma}(\int_t^{t+h}\|\boldsymbol{\phi}(t', \rvx_{t'}) - \boldsymbol{\phi}_t( \hat{\rvx}_t)\|dt')^2 \quad\text{(Jensen's)}\\
\leq& \mathbb{E}_{\gamma}(\int_t^{t+h}\|\boldsymbol{\phi}(t', \rvx_{t'}) - \boldsymbol{\phi}_t( \rvx_{t'}) + \boldsymbol{\phi}_t( \rvx_{t'}) - \boldsymbol{\phi}_t( \hat{\rvx}_t)\|dt')^2\\
\leq&\mathbb{E}_{\gamma}(\int_t^{t+h}\|\boldsymbol{\phi}(t', \rvx_{t'}) - \boldsymbol{\phi}_t( \rvx_{t'})\| + \|\boldsymbol{\phi}_t( \rvx_{t'}) - \boldsymbol{\phi}_t( \hat{\rvx}_t)\|dt')^2 \text{(Triangular)}\\
\leq&\mathbb{E}_{\gamma}(\int_t^{t+h}K_2(t'-t) + K_1\|\rvx_{t'} - \hat{\rvx}_t\|dt')^2\\
=&\mathbb{E}_{\gamma}(\frac{K_2 h^2}{2} + \int_t^{t+h}K_1\|\rvx_{t'} - \hat{\rvx}_t\|dt')^2\\
\leq&\mathbb{E}_{\gamma}(\frac{K_2 h^2}{2} + \int_t^{t+h}K_1\|\rvx_{t'}-\rvx_{t}\| + K_1\|\rvx_{t} - \hat{\rvx}_t\|dt')^2\quad\quad\text{(Triangular)}\\
=&\mathbb{E}_{\gamma}(\frac{K_2 h^2}{2} + h K_1\|\rvx_{t} - \hat{\rvx}_t\| + \int_t^{t+h}K_1\|\rvx_{t'}-\rvx_{t}\|dt')^2\\
\leq&\mathbb{E}_{\gamma}(\frac{K_2 h^2}{2} + h K_1\|\rvx_{t} - \hat{\rvx}_t\| + \int_t^{t+h}K_1K_3(t' - t)dt')^2\\
=&\mathbb{E}_{\gamma}(\frac{K_2 h^2}{2} + h K_1\|\rvx_{t} - \hat{\rvx}_t\| + \frac{K_1K_3}{2}h^2)^2. \label{eq:dis_sec_2}
\end{align}

Substituting~\eqref{eq:dis_sec_2} to~\eqref{eq:dis_sec_1},
\begin{align}
&W^2(p_{\rvx_{t+h}}, p_{\hat{\rvx}_{t+h}})\\
\leq&(1+\lambda)\mathbb{E}_{\gamma}\|\rvx_{t} - \hat{\rvx}_t\|^2 + (1+\frac{1}{\lambda}) \mathbb{E}_{\gamma}(C_1 h^2 + h K_1\|\rvx_{t} - \hat{\rvx}_t\|)^2\\\nonumber
=&(1+\lambda)\mathbb{E}_{\gamma}\|\rvx_{t} - \hat{\rvx}_t\|^2\\
& + (1+\frac{1}{\lambda}) \mathbb{E}_{\gamma}(C_1^2 h^4 + h^2 K_1^2\|\rvx_{t} - \hat{\rvx}_t\|^2 + 2 C_1 h^3 K_1\|\rvx_{t} - \hat{\rvx}_t\|)\\\nonumber
=&(1+\lambda)\mathbb{E}_{\gamma}\|\rvx_{t} - \hat{\rvx}_t\|^2\\
& + (1+\frac{1}{\lambda}) (C_1^2 h^4 + h^2 K_1^2\mathbb{E}_{\gamma}\|\rvx_{t} - \hat{\rvx}_t\|^2 + 2 C_1 h^3 K_1\mathbb{E}_{\gamma}\|\rvx_{t} - \hat{\rvx}_t\|)\\\nonumber
\leq&(1+\lambda)\mathbb{E}_{\gamma}\|\rvx_{t} - \hat{\rvx}_t\|^2\\
& + (1+\frac{1}{\lambda}) (C_1^2 h^4 + h^2 K_1^2\mathbb{E}_{\gamma}\|\rvx_{t} - \hat{\rvx}_t\|^2 + 2 C_1 h^3 K_1\sqrt{\mathbb{E}_{\gamma}\|\rvx_{t} - \hat{\rvx}_t\|^2}),
\end{align}
where $C_1 = \frac{K_2}{2} + \frac{K_1K_3}{2}$. 
Choosing $\gamma$ that minimizes $\mathbb{E}_{\gamma}\|\rvx_{t}- \hat{\rvx}_t\|^2$, 
\begin{align}
&W^2(p_{\rvx_{t+h}}, p_{\hat{\rvx}_{t+h}})\\
\leq&(1+\lambda)W^2(p_{\rvx_{t}}, p_{\hat{\rvx}_t}) + (1+\frac{1}{\lambda}) (C_1^2 h^4 + h^2 K_1^2W^2(p_{\rvx_{t}}, p_{\hat{\rvx}_t}) + 2 C_1 h^3 K_1W(p_{\rvx_{t}}, p_{\hat{\rvx}_t})). 
\end{align}
Choosing $\lambda=h$,
\begin{align}
&W^2(p_{\rvx_{t+h}}, p_{\hat{\rvx}_{t+h}})\\
\leq&(1+h)W^2(p_{\rvx_{t}}, p_{\hat{\rvx}_t}) + (1+\frac{1}{h}) (C_1^2 h^4 + h^2 K_1^2W^2(p_{\rvx_{t}}, p_{\hat{\rvx}_t}) + 2 C_1 h^3 K_1W(p_{\rvx_{t}}, p_{\hat{\rvx}_t}))\\
\leq&(1+h)W^2(p_{\rvx_{t}}, p_{\hat{\rvx}_t}) + (1+\frac{1}{h}) (C_1^2 h^4 + h^2 K_1^2W^2(p_{\rvx_{t}}, p_{\hat{\rvx}_t}) + 4 C_1 h^3 K_1 K_3)\\
\leq&(1+h)W^2(p_{\rvx_{t}}, p_{\hat{\rvx}_t}) + h K_1^2W^2(p_{\rvx_{t}}, p_{\hat{\rvx}_t}) + C h^2\\\label{eq:boundW}
\leq&[1+(1+K_1^2)h]W^2(p_{\rvx_{t}}, p_{\hat{\rvx}_t}) + C h^2,
\end{align}
where $C = 2C_1^2 + 4K_1^2K_3^2 + 8C_1K_1K_3$ is a constant. The fact that $h^4<h^3<h^2$ for $0<h<1$ and $W(p_{\rvx_{t}}, p_{\hat{\rvx}_t})$ can be bounded by $2K_3$ leads to \eqref{eq:boundW}. $W(p_{\rvx_{t}}, p_{\hat{\rvx}_t}) $ is bounded because $W^2(p_{\rvx_{t}}, p_{\hat{\rvx}_t}) \leq \mathbb{E} \|\rvx_{1} - \rvx_{0} - (\hat{\rvx}_1 -\rvx_0)\|^2 \leq 2 \mathbb{E} \|\rvx_{1} - \rvx_{0}\|^2 + 2 \mathbb{E} \|\hat{\rvx}_1 -\rvx_0\|^2\leq 4K_3^2$.

We therefore have 
\begin{equation}
W^2(p_{\rvx_{t+h}}, p_{\hat{\rvx}_{t+h}})\leq[1+(1+K_1^2)h]W^2(p_{\rvx_{t}}, p_{\hat{\rvx}_t}) + C h^2,  \\
\end{equation}
and 
\begin{equation}
W^2(p_{\rvx_{t+h}}, p_{\hat{\rvx}_{t+h}}) + \frac{Ch}{1+K_1^2}\leq[1+(1+K_1^2)h][W^2(p_{\rvx_{t}}, p_{\hat{\rvx}_t}) + \frac{Ch}{1+K_1^2}]. 
\end{equation}

It can then be shown that 
\begin{equation}
W^2(p_{\rvx_{nh}}, p_{\hat{\rvx}_{nh}})+ \frac{Ch}{1+K_1^2}\leq [1+(1+K_1^2)h]^n[W^2(p_{\rvx_{0}}, p_{\hat{\rvx}_{0}})+ \frac{Ch}{1+K_1^2}] = \frac{[1+(1+K_1^2)h]^nCh}{1+K_1^2}. \\
\end{equation}

Hence, 
\begin{align}
W^2(p_{\rvx_1}, p_{\hat{\rvx}_1})&\leq h\frac{C}{1+K_1^2}[(1 + (1+K_1^2)h)^{1/h}) - 1]\\
&\leq h\frac{C}{1+K_1^2}(\exp(1+K^2_1)-1). 
\end{align}
\end{proof}

Combining~\cref{thm:continious_error} and ~\cref{thm:discretization_error}, we have our main result:

\begin{theorem}
For two flows $\boldsymbol{\phi}_t^\theta(\rvx)$ and $\boldsymbol{\phi}_t( \rvx)$ under~\cref{assumption}, the Wasserstein distance between the distribution $p_{\hat{\rvx}_1^\theta}$ of random variable $\hat{\rvx}_1^\theta$ generated by discretization $\hat{\rvx}_{(n+1)h} = \hat{\rvx}_{nh} + \boldsymbol{\phi}_{nh}^\theta(\hat{\rvx}_{nh})$ with step-size $h$, and the distribution $p_{\rvx_1}$ of $\rvx_1 = \rvx_0 + \int_0^1 \boldsymbol{\phi}_t^\theta(\rvx_t) \diff t$ is bounded:
\begin{align}\footnotesize
&W(p_{\hat{\rvx}_1^\theta}, p_{\rvx_1}) \leq \delta \sqrt{ \exp(1 + 2K_1) } + \sqrt{h}\sqrt{\frac{C(\exp(1+K^2_1)-1)}{1+K_1^2}},
\end{align}
where $C = \frac{1}{2}K_2^2 + \frac{17}{2}K_1^2K_3^2 + 5K_1K_2K_3$.
\end{theorem}
\begin{proof}
Because Wasserstein distance is a metric,
\begin{align}
W(p_{\hat{\rvx}_1^\theta}, p_{\rvx_1}) \leq W(p_{\hat{\rvx}_1^\theta}, p_{\rvx_1^\theta}) + W(p_{\rvx_1^\theta}, p_{\rvx_1}),
\end{align}
where $p_{\rvx_1^\theta}$ is the distribution of $\rvx_1^\theta$, the random variable following the gradient flow $\boldsymbol{\phi}_t^\theta(x)$.
With the first term bounded by Theorem~\ref{thm:discretization_error} and the second term bounded by Theorem~\ref{thm:continious_error}, we complete the proof.
\end{proof}

{  
\subsection{Proof of Proposition \ref{pro:consistency}}

For a given path $\hat{p}_t(\vx)$, we are essentially solving the equation \eqref{eq:pggf} restated below
\begin{equation}
    \frac{\partial \ln\hat{p}_t(\vx)}{\partial t}  + (\nabla \ln{\hat{p}_t}(\vx) + \nabla) \cdot \boldsymbol{\phi} _t(\vx) - \mathbb{E}_{\rvx \sim p_t}\left[\frac{\partial \ln\hat{p}_t(\rvx)}{\partial t}\right]=0
\end{equation}
via minimizing a quadratic loss function. Note that the solution $\boldsymbol{\phi} _t$ for the equation above may not be unique. We will next show minimizing the quadratic loss function is consistent with solving the equation. Having an infinite number of samples implies that we are studying the behavior in expectation. Note that the precise impact of finite samples is related to the issue of generalization error, which is beyond the scope of this work. Given infinite number of particles following distribution $p_t$, the loss function in Equation \eqref{eq:pggfloss} can be written as
\begin{align}
    L_t(\theta) 
    = & \mathbb{E}_{\vx \sim p_t} \left[ \left( \frac{\partial \ln\hat{p}_t(\vx)}{\partial t}  + (\nabla \ln{\hat{p}_t}(\vx) + \nabla) \cdot \boldsymbol{\phi} ^\theta_t(\vx) - \mathbb{E}_{\rvx \sim p_t}\left[\frac{\partial \ln\hat{p}_t(\rvx)}{\partial t} \right] \right )^2 \right] \\
    = & \mathbb{E}_{\vx \sim p_t} \left[ \left( (\nabla \ln{\hat{p}_t}(\vx) + \nabla) \cdot ( \boldsymbol{\phi} ^\theta_t(\vx) - \boldsymbol{\phi} _t(\vx) )\right )^2 \right],
\end{align}
where the first relation is by definition and the second relation is by $(\nabla \ln{\hat{p}_t}(\vx) + \nabla) \cdot \boldsymbol{\phi} _t(\vx) = \mathbb{E}_{\rvx \sim p_t}\left[\frac{\partial \ln\hat{p}_t(\rvx)}{\partial t}\right] - \frac{\partial \ln\hat{p}_t(\vx)}{\partial t}$.

\paragraph{Discussion of the function space} The discussion is the same for any $t \in [0, 1]$ and we omit subscript $t$ in the following. We first need to specify the vector field class (the function space to solve the PDE) that the optimization is performed in. Define the operator $(\Tc \boldsymbol{\psi})(\vx) := (\nabla \ln p(x) + \nabla ) \cdot \boldsymbol{\psi}(\vx)$, where $\boldsymbol{\psi}$ is a differentiable vector field. Let $L^2(\mu) \subset \{\Rb^d \rightarrow \Rb\}$ be a weighted $L^2$ space with measure $d \mu(x) = p(x) d x$. It is required that $ \Tc ( \boldsymbol{\phi} ^\theta - \boldsymbol{\phi} ) 
 \in L^2(\mu)$ so that the loss function exists. This condition is satisfied by choosing the vector field such that $\nabla \ln p(x)$, $\boldsymbol{\phi} ^\theta$ and $\boldsymbol{\phi}$ all lie in $\Psi := [ W^{1,4}(\mu) ]^{d}$, which is the product space of the weighted Sobelov space $W^{1, 4}(\mu) = \{f \in L^4(p_t): \frac{\partial}{\partial x_i} f(\vx) \in L^4(p_t) \} \subset L^4(\mu)$. 


 The following proposition established the desired consistency.


\begin{proposition}[Restatement of Proposition~\ref{pro:consistency}]
Under Assumption~\ref{assumption:pde}, for any $\boldsymbol{\phi}^\theta_t$ there exists a vector-field $\boldsymbol{\phi}_t$ solution to PDE \eqref{eq:pggf} that 
\begin{align}
    \Eb_{\rvx \sim p_t} [\| \boldsymbol{\phi}_t^\theta(\rvx) - \boldsymbol{\phi}_t(\rvx) \|^2] \leq K \mathbb{E}_{\rvx \sim p_t} \left[ \left( \frac{\partial \ln\hat{p}_t(\rvx)}{\partial t}  + (\nabla \ln{\hat{p}_t}(\rvx) + \nabla) \cdot \boldsymbol{\phi} ^\theta_t(\rvx) - \mathbb{E}_{\rvx \sim p_t}\left[\frac{\partial \ln\hat{p}_t(\rvx)}{\partial t} \right] \right )^2 \right] 
\end{align}
where $K > 0$ is a universal constant factor.
\end{proposition}
\begin{proof}
We omit the subscript $t$ in $p_t$, $\boldsymbol{\phi}_t$, $\boldsymbol{\phi}_t^\theta$ for simplicity, and we let $\mu$ be the measure associated with $p_t$ as $\dm\mu(\vx) = p(\vx) \dm t$. The function class $L^2(\mu)$, $L^4(\mu)$ and $W^{1,4}(\mu)$ is defined accordingly. 


Note that $(\Psi, \|\cdot \|_{\Psi})$ is a Banach space  with 
\begin{align}
    \| \boldsymbol{\psi}\|_{\Psi}^2 := \sum_{j} \int \boldsymbol{\psi}_j(\vx)^2 d\mu(\vx) + \sum_{i, j} \int [\frac{\partial}{\partial x_i} \boldsymbol{\psi}_j(\vx) ]^2 d \mu(\vx).
\end{align}
$L^2(\mu)$ is a naturally a Banach space with $\| g \|_\mu^2 = \int g(\vx)^2 d \mu(\vx)$ for any $g \in L^2(\mu)$. 

We next show that operator $\Tc: \Psi \rightarrow L^2(\mu)$ is 
\begin{align}
    (\Tc \boldsymbol{\psi})(\vx) := (\nabla \ln p(\vx) + \nabla ) \cdot \boldsymbol{\psi}(\vx), \quad \forall \boldsymbol{\psi}\in \Psi,
\end{align}
is a bounded linear operator. The linearity is straightforward. The boundedness is because that for any $\boldsymbol{\psi}\in \Psi$ with $\| \boldsymbol{\psi}\|_{\Psi} < \infty$, 
\begin{align}
    \| \Tc \boldsymbol{\psi}\|_{\mu} \leq \|\nabla \ln p \cdot \boldsymbol{\psi} \|_\mu + \|\nabla  \cdot \boldsymbol{\psi}\|_\mu < \infty,
\end{align}
where the first inequality is by triangle inequality and the second is by the fact that 
\begin{align}
    \|\nabla \ln p \cdot \boldsymbol{\psi}  \|_\mu^2 & = \int (\nabla \ln p(\vx) \cdot \boldsymbol{\psi}(\vx))^2 d \mu(\vx) \\
    & \leq \int \| \nabla \ln p(\vx) \|^2  \| \boldsymbol{\psi}(\vx)\|^2 d \mu(\vx) \\
    & \leq \sqrt{\int \| \nabla \ln p(\vx) \|^4  d \mu(\vx)} \sqrt{\int \| \boldsymbol{\psi}(\vx)\|^4 d \mu(\vx)} < \infty,
\end{align}
and the fact that $\nabla \ln p, \boldsymbol{\psi} \in \Psi = [W^{1, 4}(\mu)]^{d}$. 

Denote by $G = \{ \Tc \boldsymbol{\psi}: \boldsymbol{\psi}\in \Psi\}$ the range of the linear operator $\Tc$ and let $N^{\Tc} = \{\boldsymbol{\psi}\in \Psi: (\Tc \boldsymbol{\psi})(x) = 0, \forall x\}$ be the null space of $\Tc$. It follows that $\Tc: \Psi / N^{\Tc} \rightarrow G$ is a bijection, where $\Psi / N^{\Tc}$ is the quotient space.  To see this bijection, observe that $\Tc\boldsymbol{\psi}\not= \Tc \boldsymbol{\phi}$ if and only if $\boldsymbol{\psi}- \boldsymbol{\phi} \not\in N^{\Tc}$. 

By the bounded inverse theorem \cite{treves2016topological}, the invertible mapping $\Tc^{-1}: G \rightarrow \Psi / N^{\Tc}$ exists and is bounded. Thus there exists a constant $K>0$ that for any $\boldsymbol{\phi}^\theta$, there is a $\boldsymbol{\phi}$ which solves the PDE and
\begin{align}
    \Eb_{\rvx \sim p} [\| \boldsymbol{\phi}^\theta(\rvx) - \boldsymbol{\phi}(\rvx) \|^2] & = \inf_{\boldsymbol{\xi} \in N^\Tc} \Eb_{\rvx \sim p} [ \| \boldsymbol{\phi}^\theta(\rvx) - \boldsymbol{\phi}(\rvx) - \boldsymbol{\xi}(\rvx) \|^2] \\
    & \leq \inf_{\boldsymbol{\xi} \in N^\Tc} \| \boldsymbol{\phi}^\theta - \boldsymbol{\phi} - \boldsymbol{\xi}\|^2_{\Psi} \\
    & \leq K \| \Tc \boldsymbol{\phi}^{\theta} - \Tc \boldsymbol{\phi} \|_\mu^2,
\end{align}
which concludes the proof.\end{proof}}

\section{Experimental Details}
The experiments are performed on Nvidia Tesla T4 GPU and Intel Xeon 8352Y CPU. To reproduce the experimental results, please refer to our code in our GitHub repo: \url{https://github.com/MingzhouFan97/PGPS}. Here we briefly summarize the setup.

\subsection{Illustrative Example}~\label{app:illu} 
Illustrated as Figure~\ref{fig:LMC_target},  the target is a mixture of two uncorrelated Gaussian with a standard deviation of $0.05$ and mean of $(1, 0)$ and $(1.5, 0)$, respectively. 
The initial particles are sampled from a two-dimensional uncorrelated Gaussian distribution with zero mean and variance of $0.1$. $200$ particles are considered in this example.
\subsection{Gaussian Mixture Examples}~\label{app:gm} 
To estimate the vector field $\phi_t$ for PGPS in both experiments, we use a two-layer perceptron with 64 hidden neurons and {\tt Sigmoid} activation function. The particle step-sizes $\psi$ is set to be $\{0.5, 0.1, 0.05, 0.01\}$, the step size for LD, PFG, SVGD, and PGPS adjustment are all set to be $10^{-2}$.
\subsection{Weight Recovery}~\label{app:wr} 
The centers of the four modes are deterministically set to be $\mu_1 = [1, 0, 0, 0, 0, 0, 0, 0]$, $\mu_2 = [0, -1, 0, 0, 0, 0, 0, 0]$, $\mu_3 = [0, 0, 1, 0, 0, 0, 0, 0]$, and $\mu_4 = [0, 0, 0, -1, 0, 0, 0, 0]$. The weights are generated by performing {\tt Softmax} over samples from a 4-dimensional standard Gaussian distribution. The NN to estimate the vector field $\phi_t$ for PGPS is a two-layer perceptron with 128 hidden neurons and {\tt Sigmoid} activation function. The step size for LD, PFG, SVGD, and PGPS adjustment are all set to be $10^{-4}$.
\subsection{Bayesian Neural Network Inference}~\label{app:bnn} 
The NN to estimate the vector field $\phi_t$ for PGPS is a two-layer perceptron with 128 hidden neurons and {\tt Sigmoid} activation function. The step size for LD, PFG, SVGD, and PGPS adjustment are all set to be $10^{-1}$. The path hyperparameter $\alpha$ is selected from $\{0, 0.2, 0.4, 0.6, 0.8, 1\}$ and $\beta$ is selected from $\{0.2, 0.4, 0.6, 0.8, 1\}$.

\section{Additional Experimental Results}
\subsection{BNN inference on UCI Dataset}~\label{app:UCI} 
We report the NLL along with the ACC results for~\cref{sec:UCI} in~\cref{tab:UCINLL}. In many datasets, SVGD has the best NLL; while in none of these benchmark experiments, SVGD can achieve the best ACC. We conjecture that this is due to variance collapse that SVGD leads to particles gathering close together on the modes and in turn being `over-confident' on the prediction so that SVGD would tend to get better NLL on certain datasets but worse on ACC. Our PGPS achieves the best ACC and second-best NLL in many of the datasets.
\begin{table*}[h]
\centering
\caption{Average negative log-likelihood~(NLL) and accuracy~(ACC) on UCI datasets over five independent runs}
\label{tab:UCINLL}\vspace{-3mm}
\scalebox{0.7}{\begin{tabular}{c|cccc|cccc}
\hline
 \multicolumn{1}{c}{} & \multicolumn{4}{c}{Negative Log-Likelihood (NLL)} & \multicolumn{4}{c}{Accuracy (ACC)} \\
 \hline 
 & \textbf{PGPS}     &  SVGD &    SGLD&       PFG&      \textbf{PGPS}   &  SVGD &    SGLD&       PFG\\
 \hline
SONAR &$0.5357 \pm0.014$&$\textbf{0.5059} \pm0.010$&$0.5099 \pm0.017$&$0.5314 \pm0.011$&$\textbf{0.7981} \pm0.023$&$0.7962 \pm0.016$&$0.7942 \pm0.024$&$0.7673 \pm0.033$\\
WINEWHITE&$1.9788 \pm0.009$&$1.9905\pm0.011$&$\textbf{1.9774}\pm0.050$&$1.9898\pm0.010$&$0.4520\pm0.010$&$0.4520\pm0.010$&$\textbf{0.4831}\pm0.049$&$0.4520\pm0.010$\\
WINERED&$1.9642\pm0.012$&$1.9566\pm0.012$&$1.9502\pm0.096$&$\textbf{1.9359}\pm0.018$&$\textbf{0.5938}\pm0.018$&$0.5770\pm0.018$&$0.5107\pm0.096$&$0.5723\pm0.019$\\
AUSTRALIAN&$0.5042 \pm 0.013$&$\textbf{0.4507} \pm0.006$&$0.5732 \pm0.161$&$0.4511 \pm0.007$&$0.8620 \pm0.009$&$0.8626 \pm0.006$&$0.7362 \pm0.157$&$\textbf{0.8643} \pm0.006$\\
HEART&$\textbf{0.9428} \pm0.030$&$1.0800 \pm0.027$&$1.0686 \pm0.131$&$1.0914 \pm0.033$&$\textbf{0.2556} \pm0.142$&$0.1801 \pm0.042$&$0.2384 \pm0.135$&$0.1762 \pm0.033$\\
GLASS&$1.6853 \pm0.030$&$\textbf{1.6664}\pm0.027$&$1.7083\pm0.145$&$1.7162\pm0.029$&$\textbf{0.5850}\pm0.080$&$0.5383\pm0.076$&$0.4561\pm0.152$&$0.4505\pm0.071$\\
COVERTYPE&$1.6016 \pm 0.014$&$\textbf{1.5981} \pm0.018$&$1.6439 \pm0.082$&$1.6241 \pm0.011$&$\textbf{0.5899} \pm0.095$&$0.4867 \pm0.006$&$0.5221 \pm0.084$&$0.5088 \pm0.053$\\
 \hline
\end{tabular}}
\end{table*}

\end{document}